\documentclass[twoside]{article}
\usepackage{openwork}
\title{Learning Robust Visual Features in Computed Tomography Enables Efficient Transfer Learning for Clinical Tasks}

\author[1,2]{Rubén Moreno-Aguado}
\author[3,4,5]{Alba Magallón}
\author[3,4,5,6]{Victor Moreno}
\author[1,7,8,9]{Yingying Fang\,\textsuperscript{$\dagger$}}
\author[1,7,8,9]{Guang Yang\,\textsuperscript{$\dagger$}}

\affil[1]{Bioengineering Department and Imperial-X, Imperial College London, London, UK}
\affil[2]{Department of Computer Science, University of Manchester, Manchester, UK}
\affil[3]{Oncology Data Analytics Program, Catalan Institute of Oncology, Barcelona, Spain}
\affil[4]{Colorectal Cancer Group, ONCOBELL Program, Institut d’Investigació Biomèdica de Bellvitge, Barcelona, Spain}
\affil[5]{Consortium for Biomedical Research in Epidemiology and Public Health, Madrid, Spain}
\affil[6]{Department of Clinical Sciences, Faculty of Medicine and Health Sciences, Universitat de Barcelona Institute of Complex Systems, University of Barcelona, Barcelona, Spain}
\affil[7]{National Heart and Lung Institute, Imperial College London, London, UK}
\affil[8]{Cardiovascular Research Centre, Royal Brompton Hospital, London, UK}
\affil[9]{School of Biomedical Engineering \& Imaging Sciences, King's College London, London, UK}

\affil[$\dagger$]{Co-corresponding authors.}

\date{}

\begin{document}
\maketitle

\begin{abstract}
\sloppy
There is substantial interest in developing artificial intelligence systems to support radiologists across tasks ranging from segmentation to report generation. Existing computed tomography (CT) foundation models have largely focused on building generalist vision-language systems capable of tasks such as question answering and report generation. However, training reliable vision-language systems requires paired image-text data at a scale that remains unavailable in computed tomography. Moreover, adapting the underlying visual representations to downstream tasks typically requires partial or full backbone fine-tuning, a computationally demanding process inaccessible to many research groups. Instead, foundation models should prioritise learning robust visual representations that enable efficient transfer to new tasks with minimal labelled data and without backbone fine-tuning. We present VoxelFM, a 3D CT foundation model trained with self-distillation using the DINO framework, which learns semantically rich features without language supervision. Pre-training used over 137,000 CT studies spanning diverse institutions, scanners, pathologies, and body regions. We evaluated VoxelFM across seven categories of clinically relevant downstream tasks using frozen backbone representations with lightweight probes: classification, regression, survival analysis, instance retrieval, localisation, segmentation, and report generation. VoxelFM matched or outperformed four existing CT foundation models across all task categories. Despite receiving no language supervision during pre-training, VoxelFM surpassed models explicitly trained with language-alignment objectives, including on report generation. In survival analysis, it was the only model to predict survival above chance, while all baselines performed at or below chance. Our results indicate that current CT foundation models perform significantly better as feature extractors for lightweight probes rather than as vision encoders for vision-language models. Model weights and training code are publicly available.
\end{abstract}
{\textbf{Keywords:} Computed tomography, Self-supervised learning, Foundation models \par}

\section{Introduction}
Modern clinical medicine is increasingly dependent on computed tomography (CT) imaging for diagnosing and monitoring a wide range of conditions, including stroke, vascular diseases, cancer, trauma, acute abdominal pain, and diffuse lung diseases \cite{rubinComputedTomographyRevolutionizing2014}. This increased reliance on CT imaging has placed substantial pressure on radiology services worldwide \cite{brulsWorkloadRadiologistsOncall2020}. Many radiology departments report longer working hours and increased fatigue, which can impair visual search, lead to interpretative errors and increase the risk of missed diagnoses \cite{hannaEffectShiftSchedule2018, alexanderMandatingLimitsWorkload2022}. Consequently, there is substantial interest in developing artificial intelligence (AI) systems that can support radiologists with tasks ranging from segmentation to report generation. Tools such as TotalSegmentator, used for organ and tumour segmentation, are already proving useful in clinical practice \cite{wasserthalTotalSegmentatorRobustSegmentation2023}. However, there is less evidence for the readiness of AI in more open-ended tasks such as multi-abnormality detection and automated report generation \cite{ryuVisionlanguageFoundationModels2025,chengRoleArtificialIntelligencebased2025}. 

Several foundation models have recently been proposed as generalists, capable of autonomously interpreting CT scans \cite{paiVisionFoundationModels2025, blankemeierMerlinComputedTomography2026, baiM3DAdvancing3D2024, wuGeneralistFoundationModel2025, hamamciDevelopingGeneralistFoundation2024, zhuangMiMMaskMask2025}. General vision-language models have already demonstrated a significant level of image understanding \cite{wangInternVL35AdvancingOpenSource2025,baiQwen3VLTechnicalReport2025}, and this could in principle be applied to make specialised vision-language models for CT, aiming to perform automatic report generation. However, it remains to be seen whether these systems can be applied reliably in clinical practice. Previous CT foundation models applied to report generation have performed worse than the same underlying vision encoder trained for detecting a specific disease or lesion \cite{hamamciDevelopingGeneralistFoundation2024}. Many existing CT foundation models make design choices that are geared towards vision-language alignment rather than robust feature extraction. They are often pre-trained using language supervision frameworks such as contrastive language-image pre-training (CLIP) \cite{210300020LearningTransferable}, as this aims to align the learned features to clinically meaningful features. While language-supervised pre-training has produced some of the strongest general-purpose vision models in natural imaging \cite{tschannenSigLIP2Multilingual2025,wangInternVL35AdvancingOpenSource2025,bolyaPerceptionEncoderBest2025}, performance scales with the availability of large volumes of paired image-text data \cite{fanScalingLanguageFreeVisual2025}. In computed tomography, where such data remain comparatively scarce, language supervision may not be the most data-efficient learning scheme. We hypothesise that the current available paired CT and radiology report data is insufficient to supervise a vision-language model in radiology. Therefore, we suggest that the focus of foundation models for radiology should be to enable efficient transfer of learned representations to new tasks with limited labelled data. 

Self-supervised approaches that learn directly from image structure may offer advantages over language supervision when less data is available. Self-distillation methods such as DINO \cite{caronEmergingPropertiesSelfSupervised2021,oquabDINOv2LearningRobust2024,simeoniDINOv32025} are particularly powerful. DINO learns representations that are useful even without fine-tuning. Caron et al. \cite{caronEmergingPropertiesSelfSupervised2021} demonstrate this by using only k-nearest-neighbour classification and linear probes in their evaluations, showing that the learned features are semantically meaningful without any task-specific adaptation of the backbone. Transfer learning with no fine-tuning would be a considerable advantage over most existing CT foundation models, which require partial or full fine-tuning of the backbone weights \cite{paiVisionFoundationModels2025, blankemeierMerlinComputedTomography2026, baiM3DAdvancing3D2024, wuGeneralistFoundationModel2025, hamamciDevelopingGeneralistFoundation2024, zhuangMiMMaskMask2025}, a computationally demanding process that is inaccessible to many research groups.

We present VoxelFM, a 3D CT foundation model based on the DINO framework. Our contributions are as follows. First, we gathered open datasets spanning various institutions, CT scanners, pathologies, and body parts. Second, we trained a 3D vision encoder using rotary positional encodings, and augmented the CT scans in size and aspect ratio, removing hard constraints on the input dimensions. Third, we evaluated across seven categories of clinically relevant tasks and demonstrate competitive or superior performance compared to existing foundation models, particularly under limited data availability. Our representations generalise across diverse clinical tasks without backbone fine-tuning and with reduced computational costs. The global class-token representations allow for volume-level tasks and the patch-level tokens for fine-grained localisation and segmentation. All pre-trained weights and training code are publicly released. 

\section{Results}
\begin{figure}[ht]
    \centering
    \includegraphics[width=1.0\linewidth]{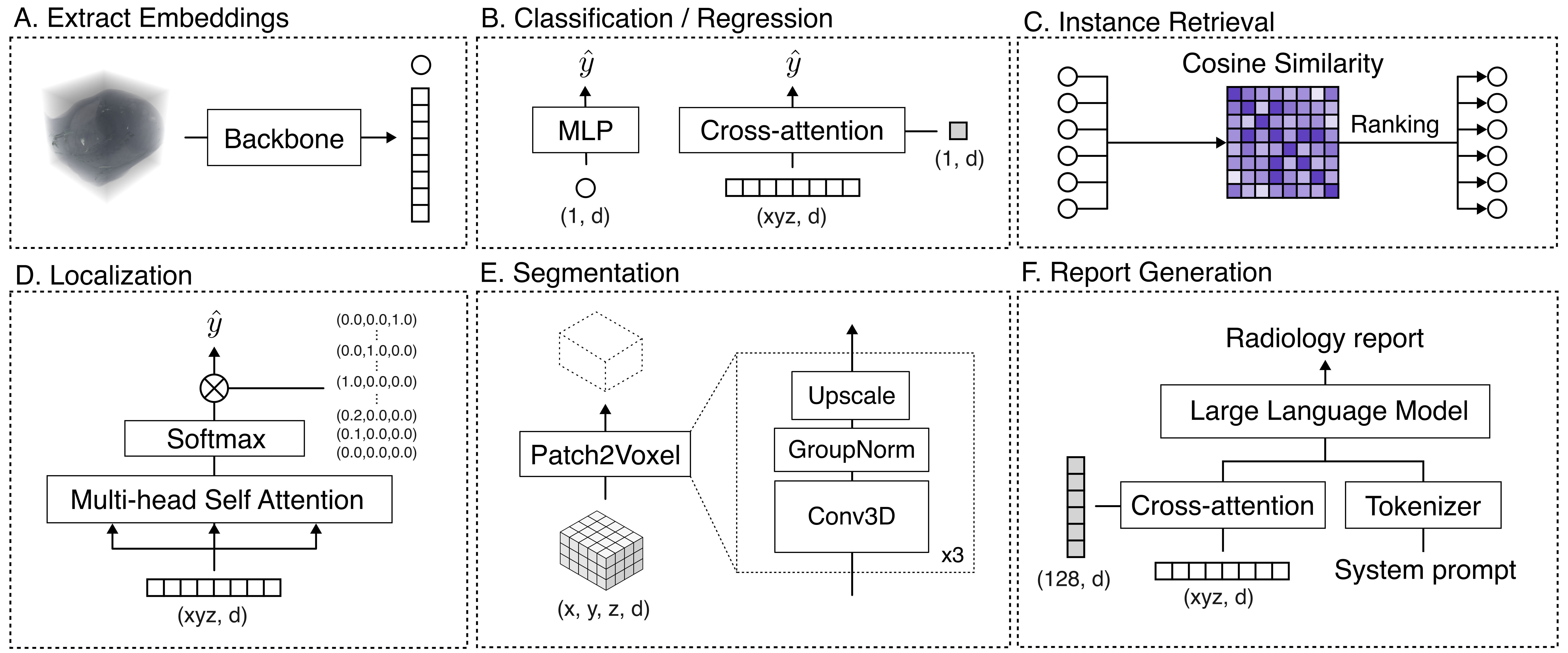}
    \caption{\textbf{Overview of the evaluation protocol.} All evaluations use pre-computed embeddings from a pre-trained encoder backbone. The first common step is (A): embeddings are extracted by the backbone and cached for downstream use. (B) Classification and regression: two methods are available depending on the token type used. For the class token, a two-layer MLP is applied, and for patch tokens, a Q-Former with cross-attention is used instead. (C) Instance retrieval: class tokens are ranked against a query token by cosine similarity. (D) Localisation: patch tokens are decoded to normalised 3D coordinates via multi-head self-attention, with a softmax operation used to weight the positional coordinates. (E) Segmentation: patch tokens are reshaped into a 3D feature map and decoded through 3D convolutional and upsampling layers. (F) Report generation: patch tokens are projected via cross-attention and combined with a system prompt before being passed to a large language model (LLM).}
    \label{fig:evalmethods}
\end{figure}

\begin{figure}[htbp]
    \centering
    \includegraphics[width=1.0\linewidth]{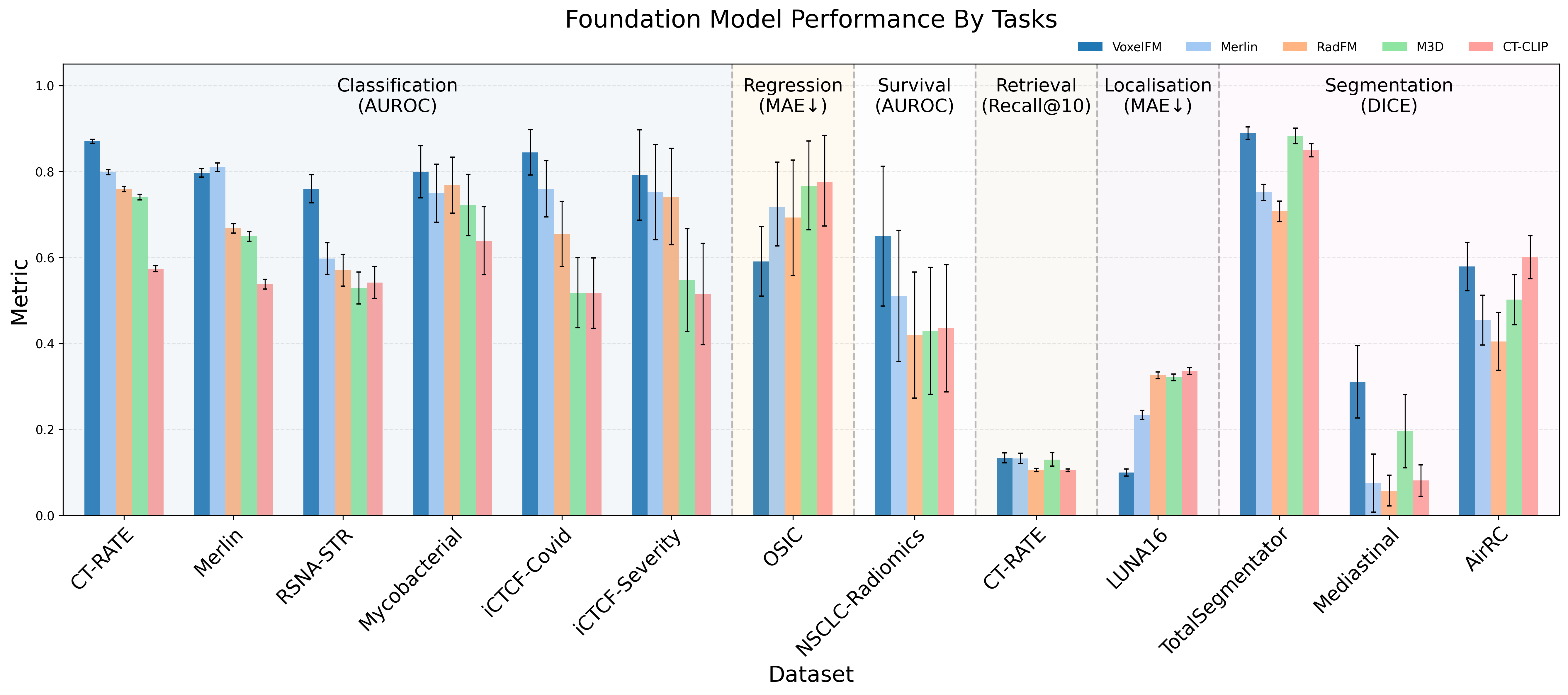}
    \caption{\textbf{Bar chart summary of model performance across six tasks.} Results are shown for classification (AUROC), regression (MAE~$\downarrow$), survival analysis (AUROC), localisation (MAE~$\downarrow$), segmentation (DICE), and retrieval (Recall@10), with 95\% confidence intervals displayed for each model. TotalSegmentator segmentation is reported as micro-averaged DICE. CT-RATE and Merlin classification results are macro-averaged over all abnormality classes. See Table~\ref{tab:results_main} for full numerical results.}
    \label{fig:mainfig}
\end{figure}

We evaluated VoxelFM across seven categories of downstream tasks with clinical interest: classification, regression, survival analysis, localisation, segmentation, report generation, and instance retrieval (Figure~\ref{fig:evalmethods}). To test whether the pre-trained representations alone can support diverse clinical applications, we froze the backbone weights in all experiments and trained only lightweight probes. VoxelFM achieved the strongest overall performance across these tasks, matching or outperforming four existing 3D CT foundation models on all categories except instance retrieval, where all models performed poorly. Figure~\ref{fig:mainfig} summarises the comparison and Table~\ref{tab:results_main} reports the full numerical results. 

\begin{landscape}
\begin{table}[p]
\centering
\begin{tabular}{lllc@{\,}lcccc}
\toprule
Task & Dataset & Metric & \multicolumn{2}{c}{VoxelFM} & Merlin & RadFM & M3D & CT-CLIP \\
\midrule
\multirow{6}{*}{Classification} & CT-RATE & AUROC & \textbf{0.870 (0.006)} & $*$ & 0.798 (0.006) & 0.759 (0.002) & 0.740 (0.008) & 0.574 (0.002) \\
 & Merlin & AUROC & 0.797 (0.005) & $ $ & \textbf{0.810 (0.005)} & 0.668 (0.006) & 0.649 (0.006) & 0.538 (0.006) \\
 & RSNA-STR & AUROC & \textbf{0.760 (0.017)} & $*$ & 0.597 (0.019) & 0.570 (0.019) & 0.529 (0.019) & 0.542 (0.019) \\
 & Mycobacterial & AUROC & \textbf{0.799 (0.031)} & $ $ & 0.750 (0.035) & 0.768 (0.033) & 0.722 (0.036) & 0.639 (0.040) \\
 & iCTCF-Covid & AUROC & \textbf{0.845 (0.027)} & $*$ & 0.760 (0.033) & 0.655 (0.038) & 0.518 (0.042) & 0.517 (0.042) \\
 & iCTCF-Severity & AUROC & \textbf{0.792 (0.054)} & $ $ & 0.752 (0.057) & 0.742 (0.057) & 0.547 (0.061) & 0.515 (0.060) \\
\midrule
\multirow{1}{*}{Regression} & OSIC & MAE ↓ & \textbf{0.591 (0.041)} & $ $ & 0.718 (0.050) & 0.693 (0.069) & 0.766 (0.053) & 0.776 (0.054) \\
\midrule
\multirow{2}{*}{Survival Analysis} & NSCLC-Radiomics & AUROC & \textbf{0.650 (0.083)} & $ $ & 0.511 (0.078) & 0.420 (0.075) & 0.430 (0.075) & 0.435 (0.076) \\
 & NSCLC-Radiomics & C-index & \textbf{0.602 (0.051)} & $ $ & 0.533 (0.041) & 0.473 (0.038) & 0.465 (0.044) & 0.500 (0.047) \\
\midrule
\multirow{1}{*}{Retrieval} & CT-RATE & Recall@10 & \textbf{0.133 (0.006)} & $ $ & 0.132 (0.006) & 0.105 (0.002) & 0.130 (0.008) & 0.105 (0.002) \\
\midrule
\multirow{1}{*}{Localisation} & LUNA16 & MAE ↓ & \textbf{0.100 (0.004)} & $*$ & 0.234 (0.005) & 0.326 (0.004) & 0.321 (0.004) & 0.336 (0.004) \\
\midrule
\multirow{4}{*}{Segmentation} & TotalSeg & DICE Micro & \textbf{0.889 (0.007)} & $ $ & 0.751 (0.010) & 0.707 (0.012) & 0.883 (0.009) & 0.850 (0.008) \\
 & TotalSeg & DICE Macro & \textbf{0.594 (0.015)} & $*$ & 0.328 (0.010) & 0.260 (0.010) & 0.537 (0.018) & 0.440 (0.013) \\
 & Mediastinal & DICE Micro & \textbf{0.311 (0.043)} & $ $ & 0.075 (0.034) & 0.058 (0.018) & 0.196 (0.043) & 0.081 (0.019) \\
 & AirRC & DICE Micro & 0.579 (0.029) & $ $ & 0.454 (0.030) & 0.405 (0.034) & 0.502 (0.030) & \textbf{0.601 (0.026)} \\
\midrule
\multirow{1}{*}{Report Generation} & CT-RATE & F1 & \textbf{0.432 (0.018)} & $*$ & 0.327 (0.018) & 0.259 (0.019) & 0.270 (0.018) & 0.197 (0.015) \\
\bottomrule
\end{tabular}
\caption{\textbf{Performance comparison across seven downstream tasks.} Results are reported for classification (AUROC), regression (MAE), survival analysis (AUROC and C-index), retrieval (Recall@10), localisation (MAE), segmentation (DICE), and report generation (F1). For regression and localisation, lower MAE is better ($\downarrow$). All other metrics are higher-is-better ($\uparrow$). Classification results on CT-RATE and Merlin are macro-averaged over all abnormality classes. TotalSegmentator segmentation results are reported as both micro- and macro-averaged DICE. Values in parenthesis are standard errors. Bold values denote the best-performing model in each row. $*$ denotes that the difference in performance between VoxelFM and next best model is statistically significant (p $<$ 0.05).} 
\label{tab:results_main}
\end{table}
\end{landscape}

\subsection{Classification}
Classification tasks included detecting abnormalities, distinguishing between similar diseases, and grading disease severity. We evaluated two probing strategies. For the class token, we trained a two-layer MLP on the global volume embedding. For the patch tokens, we trained a single-layer Q-Former that performed cross-attention over all spatial tokens. As Section~\ref{sec:ablation_cls_patch} discusses further, the class token probe performed better for smaller datasets, while the patch token probe had stronger results when sufficient training data were available.

VoxelFM achieved the highest AUROC score on five of six classification benchmarks (Table~\ref{tab:results_main}), and in three of them the superiority was statistically significant (p $<$ 0.05). For the CT-RATE chest abnormality benchmark with 18 labels, we trained a separate classifier for each label and reported the macro-average (details in Supplementary Figure S1 and Tables S1-S2). VoxelFM achieved a score of 0.870, compared to 0.798 for Merlin (p = \num{9.9e-18}). On the Merlin abdominal benchmark, which uses 30 labels and is reported as a macro-average, VoxelFM scored 0.797, which is slightly below Merlin's score of 0.810 but not statistically significant (p = \num{0.067}). This is likely due to Merlin's contrastive pre-training with radiology reports, from which the authors derived the 30 abnormality labels directly (details in Supplementary Figure S2 and Table S3). The most significant improvement was observed in the RSNA-STR pulmonary embolism detection benchmark, where VoxelFM achieved an AUROC score of 0.760, compared to 0.597 for Merlin (p = \num{1.0e-10}). VoxelFM also performed best on the smaller benchmarks. The Mycobacterial dataset focuses on distinguishing between tuberculosis and non-tuberculosis mycobacterial infections, and VoxelFM achieved a score of 0.799, which is similar to other baselines (p = 0.50). For the iCTCF-Covid detection task, it achieved a score of 0.845 (p = 0.049). For iCTCF severity grading, it achieved a score of 0.792 (p = 0.61). 

\subsection{Regression}
We used the OSIC Pulmonary Fibrosis Progression dataset to predict the forced vital capacity of a patient at the time of the CT scan. We normalised the target values so that a mean absolute error (MAE) of 1.0 corresponds to an error of one standard deviation. VoxelFM achieved the lowest MAE of 0.591, compared to 0.693 for RadFM (p = 0.20).
 
\subsection{Survival Analysis}
We used the NSCLC-Radiomics dataset of non-small cell lung cancer patients to evaluate survival prediction. Due to the dataset size being only 356 CT scans, we used the class token with an MLP probe. We modelled survival using Cox proportional hazards, treating prediction as a ranking task, and optimised this objective using negative log partial likelihood. We report the concordance index for risk score ranking and the AUROC for three-year survival classification.
VoxelFM achieved a concordance index of 0.602 and a three-year survival AUROC of 0.650, outperforming all baselines. Merlin achieved a concordance index of 0.533 and an AUROC of 0.511, which is approximately equal to chance. All other models performed below chance. While the difference between VoxelFM and Merlin is not statistically significant (p = 0.30 for C-index), VoxelFM significantly outperforms random chance with C-index 0.602\,[0.502,0.702], whereas the other baselines do not. 

\subsection{Localisation}
We evaluated lesion localisation using the LUNA16 benchmark, which contains annotated lung nodule centroids derived from the LIDC-IDRI dataset. To standardise the image size and augment the dataset, we generated multi-scale crops and resized them to $112{\times}112{\times}112$ voxels. We then used a multi-head self-attention layer, followed by a softmax-weighted sum over token positions, to decode the patch tokens into normalised 3D coordinates. VoxelFM achieved a normalised MAE of 0.100, corresponding to approximately 8.4\,mm. The next-best baseline, Merlin, achieved 0.234 (19.7\,mm) (p = \num{1.2e-86}).

\subsection{Segmentation}
We evaluated segmentation on three datasets and generated multi-scale crops as in the localisation task. We used a three-layer decoder composed of 3D convolutions followed by upsampling operations to convert patch tokens into voxel predictions.
On the TotalSegmentator benchmark, which covers 117 tissue classes and is focused on the full body, VoxelFM achieved micro-averaged DICE scores of 0.889 and macro-averaged DICE scores of 0.594. The next-best model, M3D, achieved 0.883 (p = 0.59) and 0.537 (p = 0.016), respectively. The larger advantage in the macro-averaged DICE score indicates stronger performance on less frequent tissue classes. The full breakdown results for the 117 tissue classes can be found in Supplementary Table S4. 

On the Mediastinal Lymph Node dataset, VoxelFM achieved a DICE score of 0.311, compared to 0.196 for M3D (p = 0.060). On the AirRC dataset, which covers the airway lumen, airway wall, veins and arteries, VoxelFM scored 0.579, slightly below the 0.601 score achieved by CT-CLIP (p = 0.57).

\subsection{Report Generation}

We trained report generators using a modular multimodal adaptation framework based on LLaVA \cite{liuVisualInstructionTuning}, similar to the approaches used by various CT foundation models \cite{baiM3DAdvancing3D2024,hamamciDevelopingGeneralistFoundation2024,blankemeierMerlinComputedTomography2026}. 
 
\begin{figure}[t!]
    \centering
    \includegraphics[width=1.0\linewidth]{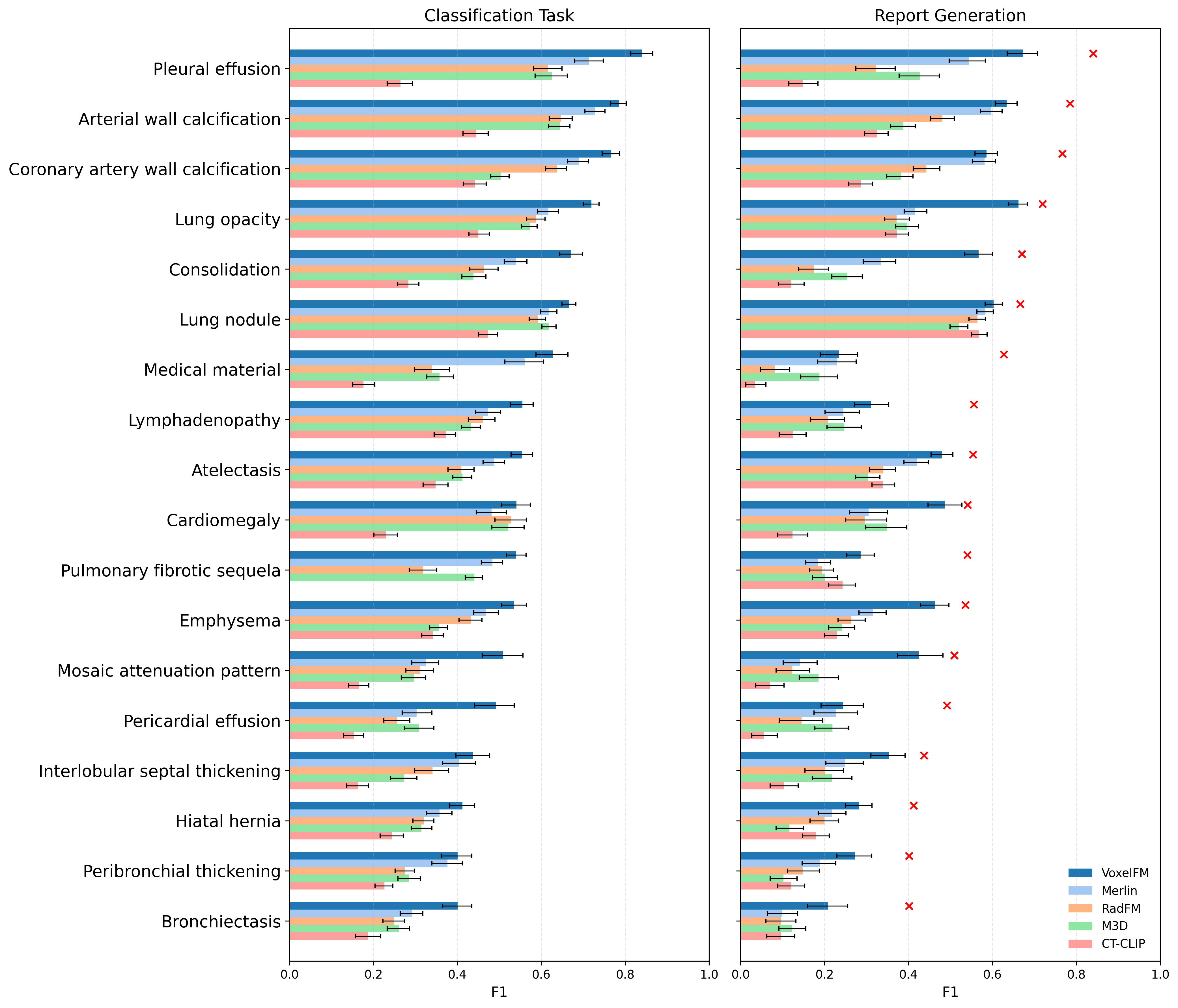}
    \caption{\textbf{Per-abnormality breakdown of classification and report generation performance across 18 findings.} (Left) Binary classification F1 scores (threshold~$= 0.5$) for each of the 18 abnormalities in the CT-RATE dataset, where an individual Q-Former probe is trained per label as described in Figure~\ref{fig:evalmethods}. (Right) Corresponding report generation F1 scores for the same 18 CT-RATE abnormality labels, aligned row-by-row with the left panel. A red cross marks the best-performing classifier result in each row.}
    \label{fig:reportgen_breakdown}
\end{figure}

VoxelFM achieved a macro-averaged F1 of 0.432, compared to 0.327 for Merlin (p = \num{5.2e-5}), 0.259 for RadFM, 0.270 for M3D, and 0.197 for CT-CLIP (details in Supplementary Figure S3 and Table S5). Despite receiving no language supervision during pre-training, VoxelFM surpassed models that were explicitly pre-trained with language-alignment objectives. For all models, report generation F1 was substantially worse than the binary classification F1 from probe classifiers trained on the same labels. Figure~\ref{fig:reportgen_breakdown} shows that the binary classifiers outperformed the report generator on every one of the 18 abnormality classes. 

\subsection{Effects of Dataset Size}
We trained probes on various fractions of available labelled training data (20\%, 40\%, 60\%, 80\%, 100\%) and evaluated them on fixed held-out test sets. We selected two tasks to represent contrasting difficulty levels: iCTCF-Covid as a relatively easy task, and RSNA-STR pulmonary embolism detection as a harder one. 

\begin{figure}
    \centering
    \includegraphics[width=1.0\linewidth]{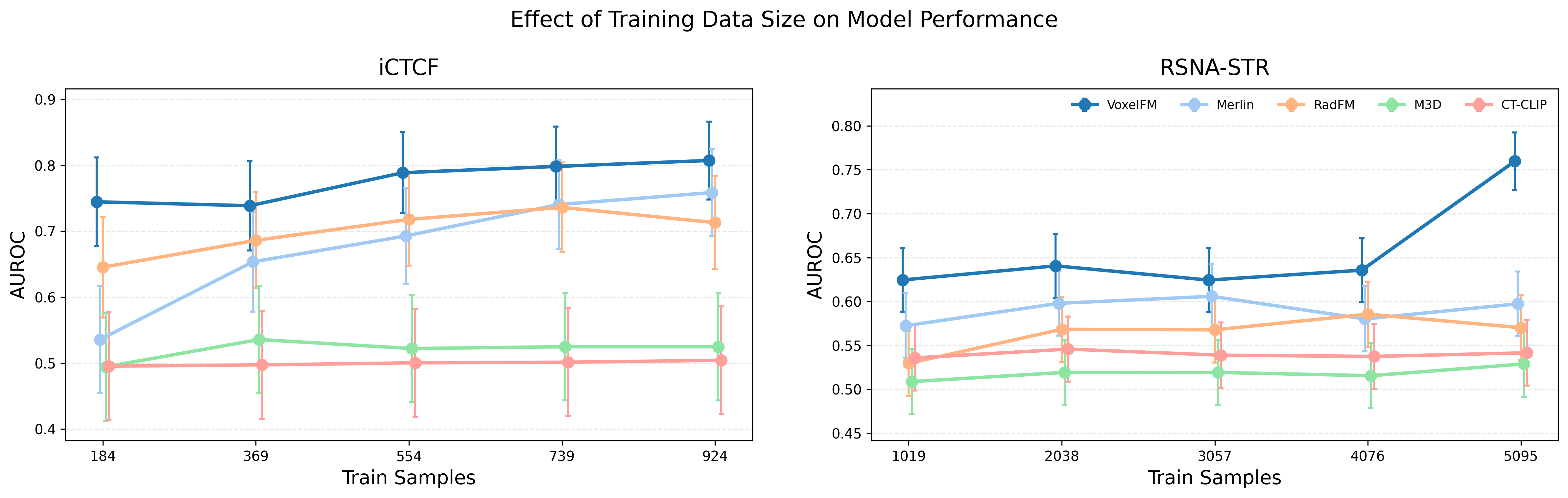}
    \caption{\textbf{Effect of downstream dataset size.} Q-Former probes are trained on various fractions of labelled training data (20\%--100\%) and evaluated on a fixed held-out test set. (Left) iCTCF-Covid. (Right) RSNA-STR. Error bars represent 95\% confidence intervals.}
    \label{fig:datasize}
\end{figure}

As shown in Figure~\ref{fig:datasize}, for iCTCF-Covid, VoxelFM achieved an AUROC of 0.74 with only 20\% of training data (184 samples), compared to 0.81 at full data. Merlin, the next-best model, dropped from 0.76 at full data to 0.54 at 20\%. For RSNA-STR, VoxelFM showed a larger absolute decline (0.63 at 20\% and 0.76 at 100\%, where 20\% corresponds to 1,019 samples). This steeper drop, even with much more available data, may reflect the greater difficulty of the task, or it may indicate that detecting pulmonary embolisms relies on patch-level features that require more training samples to learn effectively. VoxelFM remained the strongest model at every data fraction for both tasks.

\subsection{Class Token versus Patch Tokens}
\label{sec:ablation_cls_patch}
The VoxelFM backbone produces two types of output token. The class token is a single global embedding of the full volume. Patch tokens capture local spatial features across the volume. Patch tokens are required for localisation and segmentation, but either type can be used for classification tasks.
The left panel of Figure~\ref{fig:2p5d_clspatch} compares the class and patch tokens across all classification tasks, where we used an MLP for the class token case and a Q-Former for the patch token case. For the three smallest datasets (iCTCF-Covid, iCTCF-Severity, and Mycobacterial subtyping), the class token performed better. For the three larger datasets (RSNA-STR, CT-RATE, and Merlin), the patch token probe was more effective. 

\subsection{Volumetric Processing}
\begin{figure}
    \centering
    \includegraphics[width=1.0\linewidth]{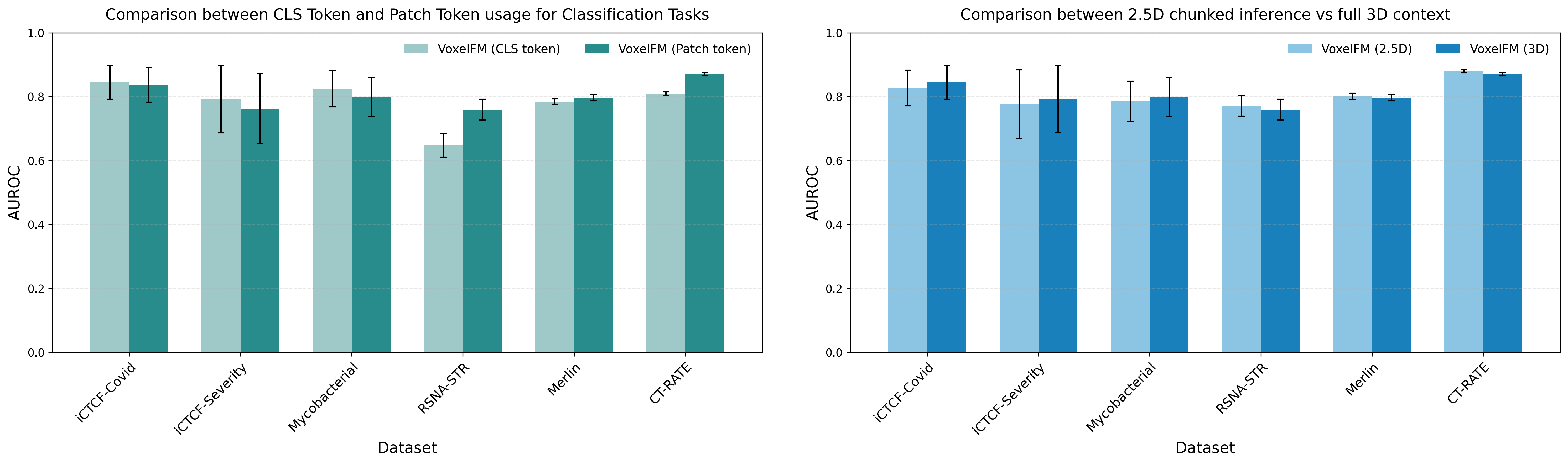}
    \caption{\textbf{Comparison of inference strategies and feature aggregation methods.} (Left) MLP applied to the class token versus a single-layer Q-Former applied to patch tokens for classification tasks. CT-RATE and Merlin results are macro-averaged over their respective abnormality labels. Error bars represent 95\% confidence intervals. (Right) Chunked 2.5D versus full 3D inference across classification tasks. }
    \label{fig:2p5d_clspatch}
\end{figure}

Some existing CT foundation models require a fixed input resolution, which makes full-volume inference impractical for large scans. We pre-trained VoxelFM with augmentations spanning a wide range of volume sizes and aspect ratios to remove this constraint. We evaluated both strategies across all classification tasks (Figure~\ref{fig:2p5d_clspatch} Right).
The AUROC scores were consistent between both inference methods for all tasks. 

\subsection{Instance Retrieval}
We evaluated instance retrieval on the CT-RATE benchmark using 18 abnormality labels. For each query scan, we constructed a retrieval set from one positive case and 99 negatives, ranked by cosine similarity against the query class token embedding. We report macro-averaged Recall@10.
All models performed near chance, with VoxelFM scoring 0.133 against a random baseline of 0.100. We discuss potential explanations for this finding in the Discussion.

\section{Discussion}
Existing foundation models for computed tomography require partial or full fine-tuning of the backbone for adaptation to clinical tasks. Caron et al.\ \cite{caronEmergingPropertiesSelfSupervised2021} demonstrated that vision encoders trained with DINO learn features robust enough to be useful without fine-tuning, relying only on linear probes or k-nearest-neighbour classification for downstream evaluation on natural images. Here, we adapted this scheme and trained a self-supervised foundation model to bring this robust feature learning to computed tomography. Our evaluation against four recent CT foundation models shows that self-supervised representations, used with frozen backbones and lightweight probes, achieve competitive or superior performance across a broad range of clinically relevant tasks. At the same time, we find that CT vision-language models tasked with report generation perform substantially worse than the same underlying vision encoder paired with a task-specific classifier. Given the current limited labelled data available, CT foundation models focused on enabling efficient transfer of learned representations to new tasks are preferable to building autonomous generalist systems.

Our evaluation includes both global and fine-grained tasks, all conducted without fine-tuning the backbone, to demonstrate that DINO-based pre-training learns robust and clinically useful embeddings. We further demonstrate how our design choices provide greater flexibility for input size and inference mode. Finally, we show that vision-language models built on CT foundation models are currently less reliable than simple probes, and we propose a corresponding change in how foundation models are applied downstream.

\subsection{Performance on Global Tasks}

Many computed tomography tasks require a single prediction for the whole volume. These tasks evaluate whether a foundation model has learned representations that encode clinically useful information, enabling linear probing to determine not only abnormalities directly present in the image, but also more abstract labels such as disease severity, presence of specific biomarkers, or prognosis.

VoxelFM achieved the highest AUROC on five of six classification benchmarks. On CT-RATE, which covers 18 predominantly pulmonary abnormality labels, VoxelFM outperformed the next-best model by nearly four points. On the Merlin benchmark, which uses 30 abdominal abnormality labels, VoxelFM scored slightly below Merlin itself, although not by a statistically significant difference. We attribute this to the fact that the 30 labels were derived directly from the radiology reports used to supervise Merlin during pre-training, giving it a direct advantage on its own evaluation set.

The most pronounced improvement was on the RSNA-STR pulmonary embolism detection benchmark. Pulmonary embolism detection is a substantially harder task than most abnormality classification benchmarks, as it may only be visible in small sections of the CT scan, and therefore not have a strong contribution on the global representation. VoxelFM also performed well on the benchmarks with smaller sample size. Mycobacterial subtyping requires distinguishing tuberculosis from non-tuberculous mycobacterial infections, and the strong result on this task supports that the learned features do not only capture whether a disease is present, but also fine-grained characteristics that differentiate between similar conditions. 

For regression, VoxelFM achieved the lowest MAE on the OSIC pulmonary fibrosis progression task, demonstrating that the learned representations can be decoded into quantitative physiological information. In survival analysis on the NSCLC-Radiomics dataset, VoxelFM was the only model to produce statistically significant risk stratification, while all other models performed at or below chance. This indicates that the self-supervised features learned information relevant for prognosis, while the language-supervised representations did not.

We investigated whether there are any important differences in using the class or patch token for classification tasks. We observed a consistent relationship between token type and dataset size. For the three smaller datasets (iCTCF-Covid, iCTCF-Severity, and Mycobacterial subtyping), the class token MLP outperformed the patch token Q-Former, whereas for the three larger datasets (RSNA-STR, CT-RATE, and Merlin), the patch token probe was superior. The patch token space is high-dimensional, and learning a good probe over it requires more training samples to avoid overfitting. The class token provides a lower-dimensional, easier-to-fit feature vector when labelled data are scarce. Therefore, we recommend using the class token probe when data is scarce and the patch token probe when sufficient labelled data is available. 

\subsection{Performance on Fine-Grained Tasks}
The DINO framework learns highly semantic patch representations in addition to the global representation, as illustrated by projecting the three principal components onto RGB channels (Supplementary Figure S4), suggesting they are well-suited to spatial tasks \cite{caronEmergingPropertiesSelfSupervised2021}. Localisation and segmentation results reveal whether the patch-level representations encode spatially precise anatomical and pathological information. 

On the LUNA16 nodule localisation task, VoxelFM substantially outperformed the next-best baseline, Merlin, representing a more than two-fold improvement in normalised MAE. 

On segmentation, VoxelFM achieved the highest scores on TotalSegmentator and Mediastinal Lymph Node benchmarks. The larger relative advantage in macro-averaged DICE on TotalSegmentator indicates stronger performance on less frequent tissue classes, where robust representations are most valuable. CT-CLIP performed best on AirRC, a dataset focused on airway and vascular structures in chest CT. As CT-CLIP was pre-trained exclusively on chest CT, the airway and vascular structures in AirRC likely fall close to its pre-training distribution, giving it an advantage on this particular benchmark, although the difference was not statistically significant. However, CT-CLIP performed substantially worse on the remaining two segmentation tasks, possibly due to limited generalisation outside of chest CTs. 

We do not intend for these segmentation results to compete with specialised models such as TotalSegmentator. Instead, they demonstrate that the representations support spatial and structural tasks, not only detection and discrimination.

\subsection{Data Efficiency}

VoxelFM showed strong performance in the low-data experiments. On iCTCF-Covid, VoxelFM maintained high AUROC scores with only 184 labelled training samples, while competing models degraded substantially. We attribute this data efficiency to two complementary reasons: the robustness of the self-supervised features, and the use of frozen backbone evaluation. Because the backbone is never updated during adaptation, only a lightweight probe head requires training, which is inherently less susceptible to overfitting on small datasets.

This property may be particularly valuable for practical applications of CT foundation models. Full backbone fine-tuning, which most existing CT foundation models require, is computationally demanding and inaccessible to many research groups. Frozen feature extraction lowers this barrier substantially. Moreover, for many clinical features of interest, obtaining large labelled datasets is infeasible or prohibitively expensive. A foundation model that can be adapted effectively with minimal labelled data addresses both of these constraints simultaneously.

\subsection{Volumetric Processing Flexibility}

A strength of our approach is that VoxelFM can be applied to diverse image sizes and resolutions without retraining. Processing the entire volume at once produces a global representation based purely on the model's feature extraction, and without applying an average pooling operation that could dilute the signal of spatially sparse features. Very large volumes impose substantial GPU memory constraints, and in such cases chunked inference may be required. Our experiments showed that classification performance was consistent between chunked and full-volume inference across all tasks, confirming that VoxelFM generalises to volumes larger than those seen during pre-training and can be used flexibly according to available computational resources.

\subsection{Limited Evidence for Report Generation}

Our results demonstrate that existing CT foundation models are not yet suitable for automatic report generation. For every one of the 18 CT-RATE abnormality classes, a simple binary classifier trained on frozen features outperformed the corresponding vision-language model on the same detection task. This finding suggests that current CT vision-language models are not yet a reliable substitute for targeted classifiers in clinical or research applications.

Report generation is a substantially harder task than classification. A classifier is trained directly on the label of interest, whereas a language model must learn to generate text that implies the correct label through much longer and noisier supervision. While general-purpose vision-language models have demonstrated proficient visual grounding in natural images, they are trained on datasets many orders of magnitude larger. Because generated reports closely adhere to the formatting conventions of the training data, they can appear deceptively plausible even when their clinical content is unreliable. We argue that task-specific classifiers remain the more reliable approach for extracting structured clinical information from CT. 

Previous studies have explored whether self-supervised or language-supervised vision encoders are more suitable as backbones for vision-language models, often concluding that language-supervised ones are preferred \cite{karamchetiPrismaticVLMsInvestigating,tongEyesWideShut2024,liuDataLanguageSupervision}. Here, despite receiving no language supervision during pre-training, VoxelFM outperformed models that were explicitly trained to align visual features with clinical text. This suggests that the quality of the visual representation is the primary bottleneck in CT vision-language models, where the scale of available data is many orders of magnitude less than what has been used to train large-scale general vision foundation models.

We believe that self-supervision provides a more efficient learning signal than language generation or alignment at the data scales typical of CT. Language supervision requires paired image-text data, which remains scarce in CT imaging, and the learning signal is constrained by the vocabulary and specificity of radiology reports. Self-supervision by self-distillation derives its signal directly from image structure and is not limited by annotation quality. Fan et al.\ \cite{fanScalingLanguageFreeVisual2025} have argued that pure visual self-supervision may exhibit better scaling behaviour than language supervision, and our results are consistent with this view. At the scales typical of CT foundation models, ranging from tens to hundreds of thousands of studies, self-supervision appears to be the more data-efficient pre-training strategy. We recognise, however, that self-supervision and language supervision have been compared extensively only at scales orders of magnitude larger than what is typical for CT. There is limited evidence on how these two approaches compare in the range of ten to five hundred thousand samples that characterises the compared baselines and others \cite{paiVisionFoundationModels2025, blankemeierMerlinComputedTomography2026, baiM3DAdvancing3D2024, wuGeneralistFoundationModel2025, hamamciDevelopingGeneralistFoundation2024, zhuangMiMMaskMask2025}. If large-scale paired CT and report data become available alongside the compute capacity to train at that scale, report generation may become viable.

\subsection{Limited Evidence for Instance Retrieval}

All models performed near chance on the instance retrieval benchmark, including VoxelFM. Global CT embeddings are influenced by many factors besides the target pathology, including scanner manufacturer, acquisition protocol, reconstruction kernel, patient demographics, and the extent of the imaged region. These factors can dominate the similarity structure of the embedding space, such that the most similar scans by cosine distance share acquisition characteristics rather than clinical findings. Based on our results, we find no evidence to support the use of embedding-based instance retrieval in clinical practice, and suggest that keyword-based searches remain more appropriate for this purpose.

\subsection{Limitations}

We acknowledge several additional limitations. First, we did not compare other baselines with full backbone fine-tuning. While we consider that the use of lightweight frozen probes is the most practically relevant evaluation for a foundation model, it means that the reported performance of competing baselines may underestimate their capacity when more labelled data and computational resources are available. Fine-tuning the backbone could potentially reduce performance differences for some models and tasks when more labelled data and compute are available. 

Second, we lack a detailed characterisation of the pre-training data distribution. Our data collection process was deliberately broad (Table~\ref{tab:training_data}), drawing from numerous public repositories, but we did not systematically analyse the distribution of scanner manufacturers, acquisition protocols, patient demographics, or pathology prevalence. This makes it difficult to predict precisely which downstream tasks VoxelFM is best suited for, or where distribution shift might degrade performance.

\subsection{Future Directions}

There are several directions in which this work could be developed. One natural next step is to combine self-distillation with language supervision as has previously been demonstrated for the DINO framework \cite{joseDINOv2MeetsText}. Increasing the size of both the pre-training dataset and the model is likely to further improve the results. Implementing the gram loss introduced in DINOv3 \cite{simeoniDINOv32025} may also improve training stability at a larger scale. Finally, as large-scale paired CT and report datasets become more widely available, it will be worthwhile revisiting the comparison between self-supervised and language-supervised pre-training.

\subsection{Conclusions}

Our evidence suggests that current CT foundation models perform significantly better as feature extractors for lightweight probes than as vision encoders for vision-language models. Reliable vision-language systems require large paired image-language datasets, which do not yet exist in CT. We suggest that a useful CT foundation model should instead focus on learning robust features that generalise regardless of CT origin. Such a model should encode clinically useful information at both the global and local level, require minimal fine-tuning for adaptation, and impose minimal constraints on input size and aspect ratio.

We addressed these requirements by basing our pre-training on self-distillation with DINO, which produces highly semantic representations as demonstrated by strong downstream performance without backbone fine-tuning. Rotary positional encodings and augmentations over image size and aspect ratio make the representations robust to variation in input dimensions. We also showed that the model can encode CT volumes either in chunked pieces, to minimise memory requirements, or as full volumes to use full 3D context, with both strategies having comparable results.

VoxelFM can be applied to a diverse set of clinically relevant tasks, from detecting the presence of diseases and lesions to fine-grained localisation and tissue segmentation, all without fine-tuning the backbone. Patch-level representations improved the performance on more difficult tasks when sufficient training data were available, while strong performance on easier tasks could be achieved with as few as 200 labelled samples. We release all pre-trained weights and training code to support future research.

\section{Methods}

\subsection{Dataset preparation}
Pre-training used more than 137,000 de-identified CT studies from the head and neck, thorax, and abdomen. Data came from large public datasets (CT-RATE, Merlin, NLST) and several smaller collections from the Cancer Imaging Archive (TCIA) and other public medical image repositories. Table~\ref{tab:training_data} lists all datasets and sample counts. Because CT-RATE and Merlin were also used for downstream evaluation, only their official training splits were used for pre-training. Validation and test splits were excluded to prevent data leakage. TCIA datasets used only for evaluation were fully excluded from pre-training.

We resampled all CT to isotropic voxel spacing using trilinear interpolation, retaining the smallest spacing dimension and limiting the largest volume side length to 768 voxels. We clipped the voxel intensities to the range $-1000$ to $1900$~HU and normalised them using global z-score statistics computed from the entire pre-training dataset. We removed non-patient background regions (air, table, and other empty areas) to reduce memory and compute cost.

\begin{figure}[t!]
    \centering
    \includegraphics[width=1.0\linewidth]{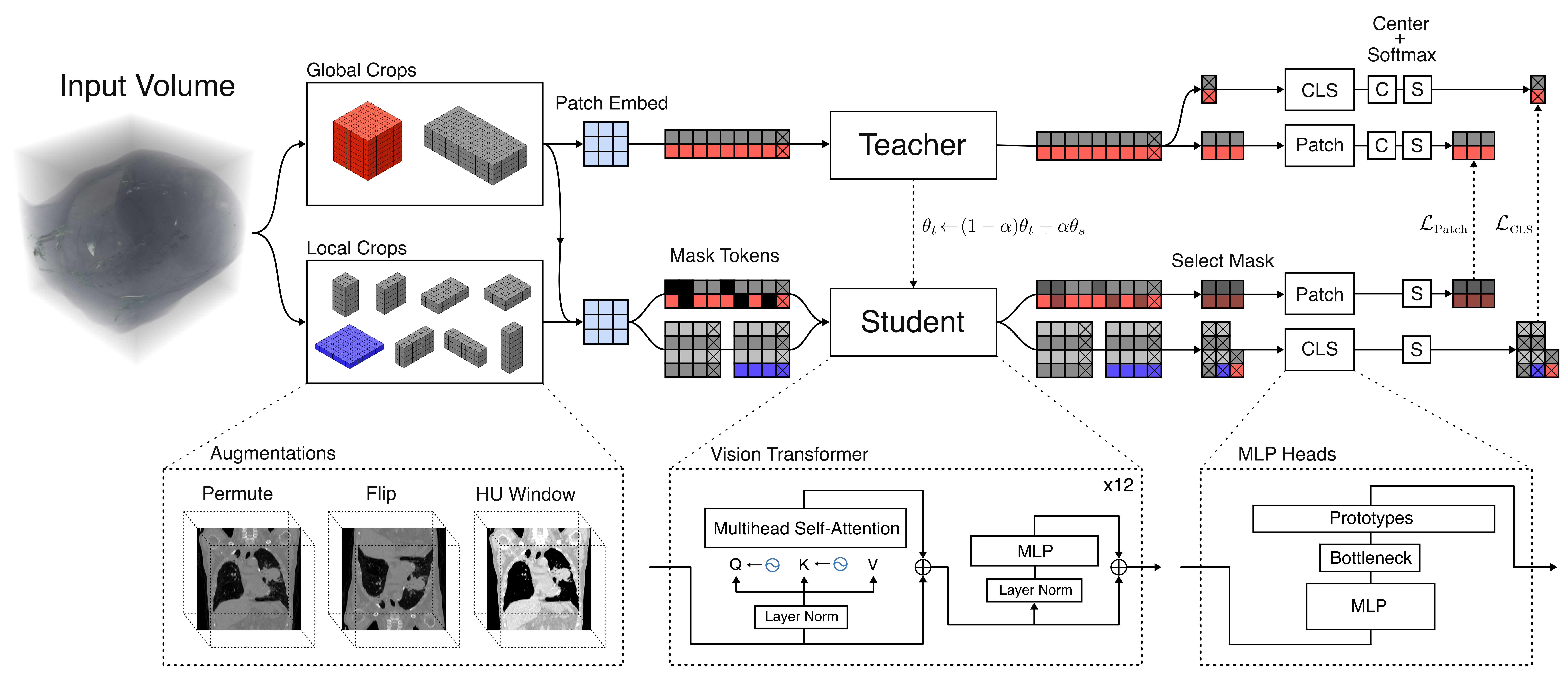}
    \caption{\textbf{Overview of the VoxelFM pre-training framework.} Student and teacher networks share identical ViT-based architectures with class and patch token heads. The teacher network is updated via an exponential moving average of the student parameters. For each CT volume, two global and eight local crops are generated. Only the student processes local crops, and random masking is applied to its global crops. Training minimises a DINO loss ($\mathcal{L}_\text{CLS}$) on class tokens and an iBOT loss ($\mathcal{L}_\text{Patch}$) on masked patch tokens, both using cross-entropy against teacher outputs. Both output heads consist of an MLP with a bottleneck layer followed by a prototype layer. The CT render is from NSCLC-Radiomics~\cite{aertsh.j.w.l.NSCLCRadiomics2014}.}
    \label{fig:methods_dino}
\end{figure}

\subsection{Pre-training Strategy}

Our pre-training strategy is illustrated in Figure~\ref{fig:methods_dino}. We adapted the methods from the DINO framework for 3D CT scans. Each sample generated two global crops covering 30–100\% of the volume and eight local crops covering 5–30\%. Global crops were processed by both the student and teacher networks. Local crops were processed only by the student.

We combined three learning objectives. The DINO objective aligned the projected class token views from all student views to the two teacher global views. The iBOT objective masked 50\% of patch tokens in the student's global views, and trained the student to predict the teacher's patch-token distribution at masked positions. The KoLeo objective encourages diversity in learned features by maximising the distance between each image's embedding and its nearest neighbours within a batch, preventing representation collapse. In addition, the projected class and patch tokens for the teacher are centred using an exponential moving average of the projections to prevent collapse to one mode, and sharpened using softmax with temperature 0.07 to prevent collapse to a uniform distribution \cite{oquabDINOv2LearningRobust2024,caronEmergingPropertiesSelfSupervised2021}.

Each view received independent augmentations: random HU window perturbations based on interquartile statistics, random axis-aligned flips, random 3D crops within the scale ranges defined above, and random permutation of slice order. Teacher weights update as an exponential moving average of student weights.

\subsubsection{Model Architecture}

The backbone was a 3D vision transformer with a $14{\times}14{\times}14$ voxel patch size, an embedding dimension of 864, twelve transformer blocks, twelve attention heads per block, an MLP expansion ratio of four, and a pre-norm design. We used 3D rotary positional embeddings. Four register tokens were used to improve stability of global representations, following Darcet et al. \cite{darcetVisionTransformersNeed2024}. Two projection heads consisted of a three-layer MLP with a hidden dimension of 2048, followed by a 256-dimensional bottleneck and output layer with 65,536 prototypes.

This model architecture was the largest feasible with our hardware (GPUs with 48GB RAM).

\subsubsection{Training Setup}

Training used distributed data parallelism across eight L40S GPUs. Each GPU processed sixteen samples per step and accumulated gradients over sixteen iterations, giving an effective batch size of 2048. Pre-training ran for 100,000 iterations. 

\begin{table}
\begin{tabularx}{\textwidth}{p{5cm} p{3.5cm} p{5cm} >{\raggedleft\arraybackslash}X}
    \toprule
    \textbf{Dataset} & \textbf{Location} & \textbf{Clinical Focus} & \textbf{Studies} \\
    \midrule

    CT-RATE \cite{hamamciDevelopingGeneralistFoundation2024}
    & Lung
    & Lung Abnormalities
    & 25,692 \\ 

    INSPECT \cite{huangINSPECTMultimodalDataset2023}
    & Lung
    & Pulmonary Embolism
    & 23,203 \\

    NLST \cite{nationallungscreeningtrialresearchteamDataNationalLung2013}
    & Lung
    & Lung Cancer
    & 13,587 \\ 

    RIDER-Lung-PET-CT \cite{muziDataRIDERLung2015}
    & Lung
    & Lung Cancer
    & 413 \\

    Lung-PET-CT-Dx \cite{liLargeScaleCTPET2020}
    & Lung
    & Lung Cancer
    & 268 \\

    Anti-PD-Lung \cite{madhaviDataAntiPD1Immunotherapy2019}
    & Lung
    & Lung Cancer
    & 211 \\

    COVID-19-NY-SBU \cite{saltzStonyBrookUniversity2021}
    & Lung
    & COVID-19
    & 101 \\

    Merlin \cite{blankemeierMerlinComputedTomography2026}
    & Abdomen
    & Abdominal Abnormalities
    & 15,331 \\

    RSNA-ABT (RATIC) \cite{rudieRSNAAbdominalTraumatic2024}
    & Abdomen
    & Traumatic Abdominal Injuries
    & 4,274 \\

    CT-Colonography \cite{smithDataCTCOLONOGRAPHY2015}
    & Abdomen
    & Colon Cancer
    & 1,707 \\

    AbdomenCT-1K \cite{maAbdomenCT1KAbdominalOrgan2022}
    & Abdomen
    & Abdominal Organ Segmentation
    & 1,061 \\

    VinDr \cite{daoPhaseRecognitionContrastenhanced2022}
    & Abdomen
    & Phase Recognition
    & 461 \\

    HCC-TACE-Seg \cite{moawadMultimodalityAnnotatedHCC2021}
    & Abdomen (Liver)
    & Hepatocellular carcinoma
    & 443 \\

    TCGA-KIRC \cite{akinCancerGenomeAtlas2016}
    & Abdomen (Kidney)
    & Renal Clear Cell Carcinoma
    & 373 \\

    StageII-Colorectal-CT \cite{tongAbdominalPelvicEnhanced2022}
    & Abdomen
    & Colorectal Cancer
    & 230 \\

    TCGA-BLCA \cite{kirkCancerGenomeAtlas2016}
    & Abdomen (Bladder)
    & Bladder Endothelial Carcinoma
    & 203 \\

    TCGA-LIHC \cite{ericksonCancerGenomeAtlas2016a}
    & Abdomen (Liver)
    & Liver Hepatocellular Carcinoma
    & 184 \\

    C4KC-KiTS \cite{hellerC4KCKiTSChallenge2019}
    & Abdomen (Kidney)
    & Kidney Cancer
    & 176 \\

    TCGA-STAD \cite{lucchesiCancerGenomeAtlas2016}
    & Abdomen (Stomach)
    & Stomach Adenocarcinoma
    & 161 \\

    TCGA-UCEC \cite{ericksonCancerGenomeAtlas2016}
    & Abdomen (Uterus)
    & Uterine Carcinoma
    & 121 \\

    CPTAC-CCRCC \cite{nationalcancerinstituteclinicalproteomictumoranalysisconsortiumcptacClinicalProteomicTumor2018a}
    & Abdomen (Kidney)
    & Clear Cell Carcinoma
    & 114 \\

    RADCURE \cite{welchRADCUREOpensourceHead2024}
    & Head \& Neck
    & Oropharyngeal Cancer
    & 3,346 \\

    HNSCC \cite{grossbergHNSCCVersion42020}
    & Head \& Neck
    & Squamous Cell Carcinoma
    & 1,171 \\

    TCGA-HNSC \cite{zuleyCancerGenomeAtlas2016}
    & Head \& Neck
    & Squamous Cell Carcinoma
    & 958 \\

    CPTAC-HNSCC \cite{nationalcancerinstituteclinicalproteomictumoranalysisconsortiumcptacClinicalProteomicTumor2018}
    & Head \& Neck
    & Head and Neck Cancer
    & 869 \\

    HNSCC-FDG-PET/CT \cite{kinahanDataACRIN6685}
    & Head \& Neck
    & Squamous Cell Carcinoma
    & 737 \\

    QIN-HEADNECK \cite{beichelDataQINHEADNECK2015}
    & Head \& Neck
    & Carcinoma
    & 480 \\

    Head-Neck-PET-CT \cite{vallieresDataHeadNeckPETCT2017}
    & Head \& Neck
    & Head and Neck Cancer
    & 325 \\

    GLIS-RT \cite{shusharinaGliomaImageSegmentation2021}
    & Head \& Neck
    & Gliomas
    & 227 \\

    HNC-IMRT-70-33 \cite{buattiCTRTSTRUCTRTDOSERTPLANSetsHead2024}
    & Head \& Neck
    & Head and Neck Cancer	
    & 209 \\
    
    HN-Cetuximab \cite{boschHeadNeckCetuximab2015}
    & Head \& Neck
    & Head and Neck Carcinomas
    & 203 \\

    Burdenko-GBM-P \cite{zolotovaBurdenkosGlioblastomaProgression2023}
    & Head \& Neck
    & Glioblastoma
    & 166 \\

    HN-RADIOMICS-HN1 \cite{weeDataHEADNECKRADIOMICSHN12019}
    & Head \& Neck
    & Head and Neck Cancer
    & 135 \\

    TotalSegmentator \cite{wasserthalTotalSegmentatorRobustSegmentation2023}
    & Whole body
    & Organ Segmentation
    & 1,092 \\

    \midrule
    \textbf{Total} & & & 137,107 \\
    \bottomrule
\end{tabularx}
\caption{\textbf{Datasets used for training the foundation model.}}
\label{tab:training_data}
\end{table}

\vspace{1cm}
\begin{table}
\begin{tabularx}{\textwidth}{p{4.5cm} X >{\raggedleft\arraybackslash}p{2cm}}
    \toprule
    \textbf{Dataset} & \textbf{Targets} & \textbf{Studies} \\
    \midrule

    CT-RATE \cite{hamamciDevelopingGeneralistFoundation2024}
    & Radiology Reports, 18 Abnormalities
    & 27,256 \\

    Merlin \cite{blankemeierMerlinComputedTomography2026}
    & Radiology Reports, 31 Abnormalities
    & 25,494 \\

    RSNA-STR \cite{colakRSNAPulmonaryEmbolism2021}
    & Has Pulmonary Embolism
    & 7,280 \\

    iCTCF \cite{ningOpenResourceClinical2020}
    & SARS-CoV-2 nucleic acids, Morbidity
    & 1,321 \\

    Mycobacterial \cite{hanMycobacterialCTImaging2025}
    & Tuberculosis
    & 1,301 \\

    TotalSegmentator \cite{wasserthalTotalSegmentatorRobustSegmentation2023}
    & Semantic segmentation 117 classes
    & 1,204 \\

    LUNA16 \cite{armatoiiiDataLIDCIDRI2015} \cite{setioValidationComparisonCombination2017}
    & Lung Nodule Coordinates and Diameter
    & 888 \\
    
    OSIC \cite{shahinOSICPulmonaryFibrosis2020}
    & Forced Vital Capacity
    & 881 \\

    Mediastinal Lymph Node \cite{idrist.MediastinalLymphNode2024}
    & Lymph Node Segmentations
    & 513 \\

    NSCLC-Radiomics \cite{aertsh.j.w.l.NSCLCRadiomics2014}
    & Mortality
    & 356 \\

    AirRC \cite{liuCustomAnnotatedDataset2025}
    & Airways, Veins, and Arteries Segmentation
    & 254 \\

    \bottomrule
\end{tabularx}
\caption{\textbf{Datasets used for downstream evaluations.} The number of studies does not include augmentations and test samples. Due to its large size, the NLST only represents a sample of the full dataset.}
\label{tab:eval_data}
\end{table}

\subsection{Baseline Selection}

\begin{table}[htbp]
\centering
\small
\begin{tabular}{llllll}
\toprule
Model & Pre-training & Training Data & Resolution & Patch Size & Architecture \\
\midrule
RadFM & Text generation & 16M 2D + 500K 3D (multi-modal) & 64x256x256 & 4x32x32 & 3D ViT \\
Merlin & CLIP + Supervised & 15K CT & 160x224x224 & --- & ResNet152 \\
M3D & CLIP & 120K CT & 32x256x256 & 4x16x16 & 3D ViT \\
CT-CLIP & CLIP & 26K CT & 240x480x480 & 10x20x20 & 2D/3D ViT \\
\midrule
VoxelFM & Self-supervised & 137K CT & 112x112x112 & 14x14x14 & 3D ViT \\
\bottomrule
\end{tabular}
\caption{\textbf{Comparison of baseline foundation models and VoxelFM.} Overview of pre-training strategies, training data, input resolution, patch size, and backbone architecture for each model. RadFM is trained on 2D and 3D modalities (X-ray, CT, MRI, PET). VoxelFM (bottom) is our proposed model.}
\label{tab:baselines}
\end{table}

\label{sec:baseline}
We compared VoxelFM against four state-of-the-art 3D medical imaging foundation models. Their differences in architecture and training strategy are given in Table~\ref{tab:baselines}.

RadFM \cite{wuGeneralistFoundationModel2025} is trained using autoregressive text generation for vision encoder pre-training on a dataset of 16M images (15.5M 2D, 500K 3D) including multiple modalities. The model uses full 3D processing with volumes of size $64{\times}256{\times}256$ and large patch sizes of $4{\times}32{\times}32$. 

Merlin \cite{blankemeierMerlinComputedTomography2026} uses contrastive learning supervised by structured EHR diagnosis codes and unstructured radiology reports. Training used 15,331 institutional CT scans with 1.8M diagnosis codes and 6M report tokens. The ResNet152 backbone processes $160{\times}224{\times}224$ volumes.

M3D \cite{baiM3DAdvancing3D2024} applies CLIP-style contrastive learning on 120K CT image-text pairs. The 3D ViT backbone processes $32{\times}256{\times}256$ volumes and patch sizes of $4{\times}16{\times}16$.

CT-CLIP \cite{hamamciDevelopingGeneralistFoundation2024} uses a vision transformer with hybrid 2D and 3D processing where initial layers process 2.5D slices independently and later layers attend in rows along the axial direction. Trained on 25,692 chest CT scans from CT-RATE with radiology reports. They processed volumes of size $240{\times}480{\times}480$ and patch sizes of $10{\times}20{\times}20$. We acknowledge that the CT-CLIP authors perform classification using similarity to positive and negative language embeddings, which differs from our approach.

\subsection{Model Selection}
The downstream datasets were split into training, validation, and test sets. Where multiple scans were available for a given patient, we ensured that each patient was included in only one split. For Merlin, we used the split that was already provided. For CT-RATE, we used their provided 'validation' set as our test set and 20\% of their 'train' set as our validation set. 

When training a probe, we computed a validation metric on the validation set at the end of each epoch. We used AUROC for classification tasks, MAE for regression and localisation tasks, C-index for survival analysis tasks, macro-averaged DICE for segmentation tasks, and cross-entropy loss for report generation tasks. After training, we selected the model checkpoint with the best validation metric and used it to perform inference on the test set to generate the reported results. 

\subsection{Report Generation}
Our method of training report generators was based on the LLaVA framework \cite{liuVisualInstructionTuning}. 
We fine-tuned a large language model using LoRA adapters \cite{huLoRALowRankAdaptation2021} to process multimodal input patches. The new multimodal inputs consisted of word tokens and image tokens transformed by a new Q-Former layer. We used Qwen3-8B as our pre-trained large language model \cite{yangQwen3TechnicalReport2025} with LoRA adapters of rank 128 and alpha of 256. We used GPT-OSS 120B to reformat CT-RATE reports into a consistent style \cite{openaiGptoss120bGptoss20bModel2025}. We then used the same model again to classify the presence of each of the 18 CT-RATE abnormality classes in generated reports for evaluation, and computed F1 against ground-truth labels. We selected Qwen3 as our base to train our vision-language models because it was a recently released model performing well on general benchmarks at the time of this study \cite{yangQwen3TechnicalReport2025}. We were restricted to the Qwen3-8B model variant due to memory constraints in GPU memory. 

\subsection{Statistical Methods}
We computed standard errors and 95\% confidence intervals for our result metrics using the following methods. For AUROC we calculated the standard error using the Hanley-McNeil method \cite{hanleyMeaningUseArea1982}. For MAE, C-index, DICE, and F1, we resampled 10,000 prediction-level samples with replacement from the test set, calculated each metric on the resampled set, and derived 95\% confidence intervals from the 2.5th and 97.5th percentiles of the resulting distribution. For comparing metrics between baselines, we used a t-test statistic. We consider p-values below 0.05 to be statistically significant. 

\section{Funding}
This project was partly supported in part by the ERC IMI (101005122), the H2020 (952172), the MRC (MC/PC/21013), the Royal Society (IEC/NSFC/211235), the NVIDIA Academic Hardware Grant Program, the SABER project supported by Boehringer Ingelheim Ltd, NIHR Imperial Biomedical Research Centre (RDA01), the Wellcome Leap Dynamic resilience program (co-funded by Temasek Trust), UKRI guarantee funding for Horizon Europe MSCA Postdoctoral Fellowships (EP/Z002206/1), UKRI MRC Research Grant, TFS Research Grants (MR/U506710/1), Swiss National Science Foundation (Grant No. 220785), and the UKRI Future Leaders Fellowship (MR/V023799/1, UKRI2738). V.M. received funding from Instituto de Salud Carlos III (ISCIII), “Programa FORTALECE del Ministerio de Ciencia e Innovación”, through the project number FORT23/00032, and grant DTS22/007 DeepRDT and CIBERESP (CB06-02-00328). The ISCIII and the Spanish Association Against Cancer (AECC) Scientific Foundation funded the TRANSCAN-3 project TANGERINE (TRANSCAN2021-071 AC22/00021). A.M. received a Juan de la Cierva fellowship JDC2023-052616-I from the Ministerio de Ciencia, Innovación y Universidades, Spain. R.M. received funding from the Engineering and Physical Sciences Research Council (EPSRC). 

\section{Acknowledgements}
We extend our sincere gratitude to the researchers and institutions who have generously made their image datasets publicly available, enabling the training and advancement of AI models. Their commitment to open science and collaboration is invaluable to the progress of the field.
We thank CERCA Programme, Generalitat de Catalunya for institutional support. 

\section{Code and model availability}
The code used to train and evaluate VoxelFM and instructions on how to access the pre-trained weights can be found at \href{https://github.com/rmaguado/VoxelFM}{https://github.com/rmaguado/VoxelFM}. 

\section{Data availability}
All the data used in this study were obtained from public datasets as shown in Table~\ref{tab:training_data}. 

\section{Ethics statement}
This study only included anonymised CT scans released under public licenses. Some head datasets have restricted access.

\section{Author Contribution Statement}

RM conceptualised the work, implemented the code, performed the experiments and wrote the manuscript. AM contributed to data acquisition and discussion of model applications. VM contributed to data acquisition, provided computing infrastructure and funding, and provided feedback on the research direction and manuscript. GY supervised the project and provided feedback on the research direction and manuscript. YF supervised the project and provided feedback on the research direction and manuscript. All authors read and approved the final version of the manuscript.

\section{Competing interests}
The authors declare no competing interests.

\printbibliography

@article{hanleyMeaningUseArea1982,
	title = {The meaning and use of the area under a receiver operating characteristic ({ROC}) curve.},
	volume = {143},
	issn = {0033-8419, 1527-1315},
	url = {http://pubs.rsna.org/doi/10.1148/radiology.143.1.7063747},
	doi = {10.1148/radiology.143.1.7063747},
	language = {en},
	number = {1},
	urldate = {2026-04-04},
	journal = {Radiology},
	author = {Hanley, J A and McNeil, B J},
	month = apr,
	year = {1982},
	pages = {29--36},
}

@inproceedings{tongEyesWideShut2024,
	address = {Seattle, WA, USA},
	title = {Eyes {Wide} {Shut}? {Exploring} the {Visual} {Shortcomings} of {Multimodal} {LLMs}},
	copyright = {https://doi.org/10.15223/policy-029},
	isbn = {979-8-3503-5300-6},
	shorttitle = {Eyes {Wide} {Shut}?},
	url = {https://ieeexplore.ieee.org/document/10655378/},
	doi = {10.1109/CVPR52733.2024.00914},
	urldate = {2026-03-31},
	booktitle = {2024 {IEEE}/{CVF} {Conference} on {Computer} {Vision} and {Pattern} {Recognition} ({CVPR})},
	publisher = {IEEE},
	author = {Tong, Shengbang and Liu, Zhuang and Zhai, Yuexiang and Ma, Yi and LeCun, Yann and Xie, Saining},
	month = jun,
	year = {2024},
	pages = {9568--9578},
}

@article{karamchetiPrismaticVLMsInvestigating,
	title = {Prismatic {VLMs}: {Investigating} the {Design} {Space} of  {Visually}-{Conditioned} {Language} {Models}},
	language = {en},
	author = {Karamcheti, Siddharth and Nair, Suraj and Balakrishna, Ashwin and Liang, Percy and Kollar, Thomas and Sadigh, Dorsa},
}

@misc{huLoRALowRankAdaptation2021,
	title = {{LoRA}: {Low}-{Rank} {Adaptation} of {Large} {Language} {Models}},
	shorttitle = {{LoRA}},
	url = {http://arxiv.org/abs/2106.09685},
	doi = {10.48550/arXiv.2106.09685},
	urldate = {2026-03-30},
	publisher = {arXiv},
	author = {Hu, Edward J. and Shen, Yelong and Wallis, Phillip and Allen-Zhu, Zeyuan and Li, Yuanzhi and Wang, Shean and Wang, Lu and Chen, Weizhu},
	month = oct,
	year = {2021},
	note = {arXiv:2106.09685 [cs]},
}

@article{joseDINOv2MeetsText,
	title = {{DINOv2} {Meets} {Text}: {A} {Unified} {Framework} for {Image}- and {Pixel}-{Level} {Vision}-{Language} {Alignment}},
	language = {en},
	author = {Jose, Cijo and Moutakanni, Theo and Kang, Dahyun and Baldassarre, Federico and Darcet, Timothee and Xu, Hu and Li, Daniel and Szafraniec, Marc and Ramamonjisoa, Michael and Oquab, Maxime and Simeoni, Oriane and Vo, Huy V and Labatut, Patrick and Bojanowski, Piotr},
}

@article{liuVisualInstructionTuning,
	title = {Visual {Instruction} {Tuning}},
	language = {en},
	author = {Liu, Haotian and Li, Chunyuan and Wu, Qingyang and Lee, Yong Jae},
}

@article{liuDataLanguageSupervision,
	title = {Data or {Language} {Supervision}: {What} {Makes} {CLIP} {Better} than {DINO}?},
	language = {en},
	author = {Liu, Yiming and Zhang, Yuhui and Ghosh, Dhruba and Schmidt, Ludwig and Yeung-Levy, Serena},
}

@misc{bolyaPerceptionEncoderBest2025,
	title = {Perception {Encoder}: {The} best visual embeddings are not at the output of the network},
	shorttitle = {Perception {Encoder}},
	url = {http://arxiv.org/abs/2504.13181},
	doi = {10.48550/arXiv.2504.13181},
	urldate = {2026-03-19},
	publisher = {arXiv},
	author = {Bolya, Daniel and Huang, Po-Yao and Sun, Peize and Cho, Jang Hyun and Madotto, Andrea and Wei, Chen and Ma, Tengyu and Zhi, Jiale and Rajasegaran, Jathushan and Rasheed, Hanoona and Wang, Junke and Monteiro, Marco and Xu, Hu and Dong, Shiyu and Ravi, Nikhila and Li, Daniel and Dollár, Piotr and Feichtenhofer, Christoph},
	month = apr,
	year = {2025},
	note = {arXiv:2504.13181 [cs]},
}

@misc{tschannenSigLIP2Multilingual2025,
	title = {{SigLIP} 2: {Multilingual} {Vision}-{Language} {Encoders} with {Improved} {Semantic} {Understanding}, {Localization}, and {Dense} {Features}},
	shorttitle = {{SigLIP} 2},
	url = {http://arxiv.org/abs/2502.14786},
	doi = {10.48550/arXiv.2502.14786},
	urldate = {2026-03-19},
	publisher = {arXiv},
	author = {Tschannen, Michael and Gritsenko, Alexey and Wang, Xiao and Naeem, Muhammad Ferjad and Alabdulmohsin, Ibrahim and Parthasarathy, Nikhil and Evans, Talfan and Beyer, Lucas and Xia, Ye and Mustafa, Basil and Hénaff, Olivier and Harmsen, Jeremiah and Steiner, Andreas and Zhai, Xiaohua},
	month = feb,
	year = {2025},
	note = {arXiv:2502.14786 [cs]},
}

@misc{wangInternVL35AdvancingOpenSource2025,
	title = {{InternVL3}.5: {Advancing} {Open}-{Source} {Multimodal} {Models} in {Versatility}, {Reasoning}, and {Efficiency}},
	shorttitle = {{InternVL3}.5},
	url = {http://arxiv.org/abs/2508.18265},
	doi = {10.48550/arXiv.2508.18265},
	urldate = {2026-03-19},
	publisher = {arXiv},
	author = {Wang, Weiyun and Gao, Zhangwei and Gu, Lixin and Pu, Hengjun and Cui, Long and Wei, Xingguang and Liu, Zhaoyang and Jing, Linglin and Ye, Shenglong and Shao, Jie and Wang, Zhaokai and Chen, Zhe and Zhang, Hongjie and Yang, Ganlin and Wang, Haomin and Wei, Qi and Yin, Jinhui and Li, Wenhao and Cui, Erfei and Chen, Guanzhou and Ding, Zichen and Tian, Changyao and Wu, Zhenyu and Xie, Jingjing and Li, Zehao and Yang, Bowen and Duan, Yuchen and Wang, Xuehui and Hou, Zhi and Hao, Haoran and Zhang, Tianyi and Li, Songze and Zhao, Xiangyu and Duan, Haodong and Deng, Nianchen and Fu, Bin and He, Yinan and Wang, Yi and He, Conghui and Shi, Botian and He, Junjun and Xiong, Yingtong and Lv, Han and Wu, Lijun and Shao, Wenqi and Zhang, Kaipeng and Deng, Huipeng and Qi, Biqing and Ge, Jiaye and Guo, Qipeng and Zhang, Wenwei and Zhang, Songyang and Cao, Maosong and Lin, Junyao and Tang, Kexian and Gao, Jianfei and Huang, Haian and Gu, Yuzhe and Lyu, Chengqi and Tang, Huanze and Wang, Rui and Lv, Haijun and Ouyang, Wanli and Wang, Limin and Dou, Min and Zhu, Xizhou and Lu, Tong and Lin, Dahua and Dai, Jifeng and Su, Weijie and Zhou, Bowen and Chen, Kai and Qiao, Yu and Wang, Wenhai and Luo, Gen},
	month = aug,
	year = {2025},
	note = {arXiv:2508.18265 [cs]},
}

@misc{baiQwen3VLTechnicalReport2025,
	title = {Qwen3-{VL} {Technical} {Report}},
	url = {http://arxiv.org/abs/2511.21631},
	doi = {10.48550/arXiv.2511.21631},
	urldate = {2026-03-19},
	publisher = {arXiv},
	author = {Bai, Shuai and Cai, Yuxuan and Chen, Ruizhe and Chen, Keqin and Chen, Xionghui and Cheng, Zesen and Deng, Lianghao and Ding, Wei and Gao, Chang and Ge, Chunjiang and Ge, Wenbin and Guo, Zhifang and Huang, Qidong and Huang, Jie and Huang, Fei and Hui, Binyuan and Jiang, Shutong and Li, Zhaohai and Li, Mingsheng and Li, Mei and Li, Kaixin and Lin, Zicheng and Lin, Junyang and Liu, Xuejing and Liu, Jiawei and Liu, Chenglong and Liu, Yang and Liu, Dayiheng and Liu, Shixuan and Lu, Dunjie and Luo, Ruilin and Lv, Chenxu and Men, Rui and Meng, Lingchen and Ren, Xuancheng and Ren, Xingzhang and Song, Sibo and Sun, Yuchong and Tang, Jun and Tu, Jianhong and Wan, Jianqiang and Wang, Peng and Wang, Pengfei and Wang, Qiuyue and Wang, Yuxuan and Xie, Tianbao and Xu, Yiheng and Xu, Haiyang and Xu, Jin and Yang, Zhibo and Yang, Mingkun and Yang, Jianxin and Yang, An and Yu, Bowen and Zhang, Fei and Zhang, Hang and Zhang, Xi and Zheng, Bo and Zhong, Humen and Zhou, Jingren and Zhou, Fan and Zhou, Jing and Zhu, Yuanzhi and Zhu, Ke},
	month = nov,
	year = {2025},
	note = {arXiv:2511.21631 [cs]},
}

@article{chengRoleArtificialIntelligencebased2025,
	title = {The role of artificial intelligence-based foundation models and “copilots” in cancer pathology: potential and challenges},
	volume = {45},
	issn = {0392-9078},
	shorttitle = {The role of artificial intelligence-based foundation models and “copilots” in cancer pathology},
	url = {https://pmc.ncbi.nlm.nih.gov/articles/PMC12763834/},
	doi = {10.1186/s13046-025-03592-4},
	urldate = {2026-03-19},
	journal = {Journal of Experimental \& Clinical Cancer Research : CR},
	author = {Cheng, Cillian H. and Wong, Chi Chun},
	month = nov,
	year = {2025},
	pages = {2},
}

@article{ryuVisionlanguageFoundationModels2025,
	title = {Vision-language foundation models for medical imaging: a review of current practices and innovations},
	volume = {15},
	issn = {2093-985X},
	shorttitle = {Vision-language foundation models for medical imaging},
	url = {https://doi.org/10.1007/s13534-025-00484-6},
	doi = {10.1007/s13534-025-00484-6},
	language = {en},
	number = {5},
	urldate = {2026-03-19},
	journal = {Biomedical Engineering Letters},
	author = {Ryu, Ji Seung and Kang, Hyunyoung and Chu, Yuseong and Yang, Sejung},
	month = sep,
	year = {2025},
	pages = {809--830},
}

@article{rubinComputedTomographyRevolutionizing2014,
	title = {Computed {Tomography}: {Revolutionizing} the {Practice} of {Medicine} for 40                     {Years}},
	volume = {273},
	issn = {0033-8419},
	shorttitle = {Computed {Tomography}},
	url = {https://pubs.rsna.org/doi/10.1148/radiol.14141356},
	doi = {10.1148/radiol.14141356},
	number = {2S},
	urldate = {2026-03-19},
	journal = {Radiology},
	publisher = {Radiological Society of North America},
	author = {Rubin, Geoffrey D.},
	month = nov,
	year = {2014},
	pages = {S45--S74},
}

@misc{openaiGptoss120bGptoss20bModel2025,
	title = {gpt-oss-120b \& gpt-oss-20b {Model} {Card}},
	url = {http://arxiv.org/abs/2508.10925},
	doi = {10.48550/arXiv.2508.10925},
	urldate = {2026-03-07},
	publisher = {arXiv},
	author = {OpenAI and Agarwal, Sandhini and Ahmad, Lama and Ai, Jason and Altman, Sam and Applebaum, Andy and Arbus, Edwin and Arora, Rahul K. and Bai, Yu and Baker, Bowen and Bao, Haiming and Barak, Boaz and Bennett, Ally and Bertao, Tyler and Brett, Nivedita and Brevdo, Eugene and Brockman, Greg and Bubeck, Sebastien and Chang, Che and Chen, Kai and Chen, Mark and Cheung, Enoch and Clark, Aidan and Cook, Dan and Dukhan, Marat and Dvorak, Casey and Fives, Kevin and Fomenko, Vlad and Garipov, Timur and Georgiev, Kristian and Glaese, Mia and Gogineni, Tarun and Goucher, Adam and Gross, Lukas and Guzman, Katia Gil and Hallman, John and Hehir, Jackie and Heidecke, Johannes and Helyar, Alec and Hu, Haitang and Huet, Romain and Huh, Jacob and Jain, Saachi and Johnson, Zach and Koch, Chris and Kofman, Irina and Kundel, Dominik and Kwon, Jason and Kyrylov, Volodymyr and Le, Elaine Ya and Leclerc, Guillaume and Lennon, James Park and Lessans, Scott and Lezcano-Casado, Mario and Li, Yuanzhi and Li, Zhuohan and Lin, Ji and Liss, Jordan and Lily and Liu and Liu, Jiancheng and Lu, Kevin and Lu, Chris and Martinovic, Zoran and McCallum, Lindsay and McGrath, Josh and McKinney, Scott and McLaughlin, Aidan and Mei, Song and Mostovoy, Steve and Mu, Tong and Myles, Gideon and Neitz, Alexander and Nichol, Alex and Pachocki, Jakub and Paino, Alex and Palmie, Dana and Pantuliano, Ashley and Parascandolo, Giambattista and Park, Jongsoo and Pathak, Leher and Paz, Carolina and Peran, Ludovic and Pimenov, Dmitry and Pokrass, Michelle and Proehl, Elizabeth and Qiu, Huida and Raila, Gaby and Raso, Filippo and Ren, Hongyu and Richardson, Kimmy and Robinson, David and Rotsted, Bob and Salman, Hadi and Sanjeev, Suvansh and Schwarzer, Max and Sculley, D. and Sikchi, Harshit and Simon, Kendal and Singhal, Karan and Song, Yang and Stuckey, Dane and Sun, Zhiqing and Tillet, Philippe and Toizer, Sam and Tsimpourlas, Foivos and Vyas, Nikhil and Wallace, Eric and Wang, Xin and Wang, Miles and Watkins, Olivia and Weil, Kevin and Wendling, Amy and Whinnery, Kevin and Whitney, Cedric and Wong, Hannah and Yang, Lin and Yang, Yu and Yasunaga, Michihiro and Ying, Kristen and Zaremba, Wojciech and Zhan, Wenting and Zhang, Cyril and Zhang, Brian and Zhang, Eddie and Zhao, Shengjia},
	month = aug,
	year = {2025},
	note = {arXiv:2508.10925 [cs]},
}

@misc{yangQwen3TechnicalReport2025,
	title = {Qwen3 {Technical} {Report}},
	url = {http://arxiv.org/abs/2505.09388},
	doi = {10.48550/arXiv.2505.09388},
	urldate = {2026-03-07},
	publisher = {arXiv},
	author = {Yang, An and Li, Anfeng and Yang, Baosong and Zhang, Beichen and Hui, Binyuan and Zheng, Bo and Yu, Bowen and Gao, Chang and Huang, Chengen and Lv, Chenxu and Zheng, Chujie and Liu, Dayiheng and Zhou, Fan and Huang, Fei and Hu, Feng and Ge, Hao and Wei, Haoran and Lin, Huan and Tang, Jialong and Yang, Jian and Tu, Jianhong and Zhang, Jianwei and Yang, Jianxin and Yang, Jiaxi and Zhou, Jing and Zhou, Jingren and Lin, Junyang and Dang, Kai and Bao, Keqin and Yang, Kexin and Yu, Le and Deng, Lianghao and Li, Mei and Xue, Mingfeng and Li, Mingze and Zhang, Pei and Wang, Peng and Zhu, Qin and Men, Rui and Gao, Ruize and Liu, Shixuan and Luo, Shuang and Li, Tianhao and Tang, Tianyi and Yin, Wenbiao and Ren, Xingzhang and Wang, Xinyu and Zhang, Xinyu and Ren, Xuancheng and Fan, Yang and Su, Yang and Zhang, Yichang and Zhang, Yinger and Wan, Yu and Liu, Yuqiong and Wang, Zekun and Cui, Zeyu and Zhang, Zhenru and Zhou, Zhipeng and Qiu, Zihan},
	month = may,
	year = {2025},
	note = {arXiv:2505.09388 [cs]},
}

@article{blankemeierMerlinComputedTomography2026,
	title = {Merlin: a computed tomography vision–language foundation model and dataset},
	copyright = {2026 The Author(s), under exclusive licence to Springer Nature Limited},
	issn = {1476-4687},
	shorttitle = {Merlin},
	url = {https://www.nature.com/articles/s41586-026-10181-8},
	doi = {10.1038/s41586-026-10181-8},
	language = {en},
	urldate = {2026-03-05},
	journal = {Nature},
	publisher = {Nature Publishing Group},
	author = {Blankemeier, Louis and Kumar, Ashwin and Cohen, Joseph Paul and Liu, Jiaming and Liu, Longchao and Van Veen, Dave and Gardezi, Syed Jamal Safdar and Yu, Hongkun and Paschali, Magdalini and Chen, Zhihong and Delbrouck, Jean-Benoit and Reis, Eduardo and Holland, Robbie and Truyts, Cesar and Bluethgen, Christian and Wu, Yufu and Lian, Long and Jensen, Malte Engmann Kjeldskov and Ostmeier, Sophie and Varma, Maya and Valanarasu, Jeya Maria Jose and Fang, Zhongnan and Huo, Zepeng and Nabulsi, Zaid and Ardila, Diego and Weng, Wei-Hung and Junior, Edson Amaro and Ahuja, Neera and Fries, Jason and Shah, Nigam H. and Zaharchuk, Greg and Willis, Marc and Yala, Adam and Johnston, Andrew and Boutin, Robert D. and Wentland, Andrew and Langlotz, Curtis P. and Hom, Jason and Gatidis, Sergios and Chaudhari, Akshay S.},
	month = mar,
	year = {2026},
	pages = {1--11},
}

@article{liuCustomAnnotatedDataset2025,
	title = {A {Custom} {Annotated} {Dataset} for {Segmentation} of {Pulmonary} {Veins}, {Arteries}, and {Airways}},
	volume = {12},
	copyright = {2025 The Author(s)},
	issn = {2052-4463},
	url = {https://www.nature.com/articles/s41597-025-06074-6},
	doi = {10.1038/s41597-025-06074-6},
	language = {en},
	number = {1},
	urldate = {2026-02-01},
	journal = {Scientific Data},
	publisher = {Nature Publishing Group},
	author = {Liu, Jian and Zhang, Zheng and Niu, Bing and Kang, Shuai and Ren, Juan and Wang, Lei and Xu, Kai},
	month = nov,
	year = {2025},
	pages = {1806},
}

@misc{fanScalingLanguageFreeVisual2025,
	title = {Scaling {Language}-{Free} {Visual} {Representation} {Learning}},
	url = {http://arxiv.org/abs/2504.01017},
	doi = {10.48550/arXiv.2504.01017},
	urldate = {2026-02-01},
	publisher = {arXiv},
	author = {Fan, David and Tong, Shengbang and Zhu, Jiachen and Sinha, Koustuv and Liu, Zhuang and Chen, Xinlei and Rabbat, Michael and Ballas, Nicolas and LeCun, Yann and Bar, Amir and Xie, Saining},
	month = apr,
	year = {2025},
	note = {arXiv:2504.01017 [cs]},
}

@misc{darcetVisionTransformersNeed2024,
	title = {Vision {Transformers} {Need} {Registers}},
	url = {http://arxiv.org/abs/2309.16588},
	doi = {10.48550/arXiv.2309.16588},
	urldate = {2025-12-05},
	publisher = {arXiv},
	author = {Darcet, Timothée and Oquab, Maxime and Mairal, Julien and Bojanowski, Piotr},
	month = apr,
	year = {2024},
	note = {arXiv:2309.16588 [cs]},
}

@article{daoPhaseRecognitionContrastenhanced2022,
	title = {Phase recognition in contrast-enhanced {CT} scans based on deep learning and random sampling},
	volume = {49},
	copyright = {© 2022 American Association of Physicists in Medicine.},
	issn = {2473-4209},
	url = {https://onlinelibrary.wiley.com/doi/abs/10.1002/mp.15551},
	doi = {10.1002/mp.15551},
	language = {en},
	number = {7},
	urldate = {2025-11-28},
	journal = {Medical Physics},
	author = {Dao, Binh T. and Nguyen, Thang V. and Pham, Hieu H. and Nguyen, Ha Q.},
	year = {2022},
	note = {\_eprint: https://aapm.onlinelibrary.wiley.com/doi/pdf/10.1002/mp.15551},
	pages = {4518--4528},
}

@article{brulsWorkloadRadiologistsOncall2020,
	title = {Workload for radiologists during on-call hours: dramatic increase in the past 15 years},
	volume = {11},
	issn = {1869-4101},
	shorttitle = {Workload for radiologists during on-call hours},
	url = {https://doi.org/10.1186/s13244-020-00925-z},
	doi = {10.1186/s13244-020-00925-z},
	language = {en},
	number = {1},
	urldate = {2025-11-18},
	journal = {Insights into Imaging},
	author = {Bruls, R. J. M. and Kwee, R. M.},
	month = nov,
	year = {2020},
	pages = {121},
}

@misc{hamamciDevelopingGeneralistFoundation2024,
	title = {Developing {Generalist} {Foundation} {Models} from a {Multimodal} {Dataset} for {3D} {Computed} {Tomography}},
	url = {http://arxiv.org/abs/2403.17834},
	doi = {10.48550/arXiv.2403.17834},
	urldate = {2025-11-01},
	publisher = {arXiv},
	author = {Hamamci, Ibrahim Ethem and Er, Sezgin and Almas, Furkan and Simsek, Ayse Gulnihan and Esirgun, Sevval Nil and Dogan, Irem and Dasdelen, Muhammed Furkan and Durugol, Omer Faruk and Wittmann, Bastian and Amiranashvili, Tamaz and Simsar, Enis and Simsar, Mehmet and Erdemir, Emine Bensu and Alanbay, Abdullah and Sekuboyina, Anjany and Lafci, Berkan and Bluethgen, Christian and Ozdemir, Mehmet Kemal and Menze, Bjoern},
	month = oct,
	year = {2024},
	note = {arXiv:2403.17834 [cs]
version: 2},
}

@misc{caronEmergingPropertiesSelfSupervised2021,
	title = {Emerging {Properties} in {Self}-{Supervised} {Vision} {Transformers}},
	url = {http://arxiv.org/abs/2104.14294},
	doi = {10.48550/arXiv.2104.14294},
	urldate = {2025-11-01},
	publisher = {arXiv},
	author = {Caron, Mathilde and Touvron, Hugo and Misra, Ishan and Jégou, Hervé and Mairal, Julien and Bojanowski, Piotr and Joulin, Armand},
	month = may,
	year = {2021},
	note = {arXiv:2104.14294 [cs]},
}

@article{ningOpenResourceClinical2020,
	title = {Open resource of clinical data from patients with pneumonia for the prediction of {COVID}-19 outcomes via deep learning},
	volume = {4},
	issn = {2157-846X},
	url = {https://www.nature.com/articles/s41551-020-00633-5},
	doi = {10.1038/s41551-020-00633-5},
	language = {en},
	number = {12},
	urldate = {2026-02-01},
	journal = {Nature Biomedical Engineering},
	author = {Ning, Wanshan and Lei, Shijun and Yang, Jingjing and Cao, Yukun and Jiang, Peiran and Yang, Qianqian and Zhang, Jiao and Wang, Xiaobei and Chen, Fenghua and Geng, Zhi and Xiong, Liang and Zhou, Hongmei and Guo, Yaping and Zeng, Yulan and Shi, Heshui and Wang, Lin and Xue, Yu and Wang, Zheng},
	month = nov,
	year = {2020},
	pages = {1197--1207},
}

@misc{aertsh.j.w.l.NSCLCRadiomics2014,
	title = {{NSCLC}-{Radiomics}},
	doi = {10.7937/K9/TCIA.2015.PF0M9REI},
	publisher = {The Cancer Imaging Archive},
	author = {{Aerts, H.J.W.L.} and {Wee, L.} and {Rios Velazquez, E.} and {Leijenaar, R.T.H.} and {Parmar, C.} and {Carvalho, S.} and {Bussink, J.} and {Monshouwer, R.} and {Haibe-Kains, B.} and {Rietveld, D.} and {Hoebers, F.} and {Rietbergen, M. M.} and {Leemans, C. R.} and {Dekker, A.} and {Quackenbush, J.} and {Gillies, R. J.} and {Lambin, P.}},
	year = {2014},
}

@article{setioValidationComparisonCombination2017,
	title = {Validation, comparison, and combination of algorithms for automatic detection of pulmonary nodules in computed tomography images: {The} {LUNA16} challenge},
	volume = {42},
	issn = {1361-8415},
	shorttitle = {Validation, comparison, and combination of algorithms for automatic detection of pulmonary nodules in computed tomography images},
	url = {https://www.sciencedirect.com/science/article/pii/S1361841517301020},
	doi = {10.1016/j.media.2017.06.015},
	urldate = {2025-12-01},
	journal = {Medical Image Analysis},
	author = {Setio, Arnaud Arindra Adiyoso and Traverso, Alberto and de Bel, Thomas and Berens, Moira S. N. and Bogaard, Cas van den and Cerello, Piergiorgio and Chen, Hao and Dou, Qi and Fantacci, Maria Evelina and Geurts, Bram and Gugten, Robbert van der and Heng, Pheng Ann and Jansen, Bart and de Kaste, Michael M. J. and Kotov, Valentin and Lin, Jack Yu-Hung and Manders, Jeroen T. M. C. and Sóñora-Mengana, Alexander and García-Naranjo, Juan Carlos and Papavasileiou, Evgenia and Prokop, Mathias and Saletta, Marco and Schaefer-Prokop, Cornelia M and Scholten, Ernst T. and Scholten, Luuk and Snoeren, Miranda M. and Torres, Ernesto Lopez and Vandemeulebroucke, Jef and Walasek, Nicole and Zuidhof, Guido C. A. and Ginneken, Bram van and Jacobs, Colin},
	month = dec,
	year = {2017},
	pages = {1--13},
}

@misc{akinCancerGenomeAtlas2016,
	title = {The {Cancer} {Genome} {Atlas} {Kidney} {Renal} {Clear} {Cell} {Carcinoma} {Collection} ({TCGA}-{KIRC})},
	copyright = {Creative Commons Attribution 3.0 Unported},
	url = {https://www.cancerimagingarchive.net/collection/tcga-kirc/},
	doi = {10.7937/K9/TCIA.2016.V6PBVTDR},
	urldate = {2025-12-01},
	publisher = {The Cancer Imaging Archive},
	author = {Akin, Oguz and Elnajjar, Pierre and Heller, Matthew and Jarosz, Rose and Erickson, Bradley J. and Kirk, Shanah and Lee, Yueh and Linehan, Marston W. and Gautam, Rabindra and Vikram, Raghu and Garcia, Kimberly M. and Roche, Charles and Bonaccio, Ermelinda and Filippini, Joe},
	collaborator = {{TCIA Team}},
	year = {2016},
}

@misc{210300020LearningTransferable,
	title = {[2103.00020] {Learning} {Transferable} {Visual} {Models} {From} {Natural} {Language} {Supervision}},
	url = {https://arxiv.org/abs/2103.00020},
	urldate = {2025-11-23},
}

@misc{simeoniDINOv32025,
	title = {{DINOv3}},
	url = {http://arxiv.org/abs/2508.10104},
	doi = {10.48550/arXiv.2508.10104},
	urldate = {2025-11-23},
	publisher = {arXiv},
	author = {Siméoni, Oriane and Vo, Huy V. and Seitzer, Maximilian and Baldassarre, Federico and Oquab, Maxime and Jose, Cijo and Khalidov, Vasil and Szafraniec, Marc and Yi, Seungeun and Ramamonjisoa, Michaël and Massa, Francisco and Haziza, Daniel and Wehrstedt, Luca and Wang, Jianyuan and Darcet, Timothée and Moutakanni, Théo and Sentana, Leonel and Roberts, Claire and Vedaldi, Andrea and Tolan, Jamie and Brandt, John and Couprie, Camille and Mairal, Julien and Jégou, Hervé and Labatut, Patrick and Bojanowski, Piotr},
	month = aug,
	year = {2025},
	note = {arXiv:2508.10104 [cs]},
}

@misc{oquabDINOv2LearningRobust2024,
	title = {{DINOv2}: {Learning} {Robust} {Visual} {Features} without {Supervision}},
	shorttitle = {{DINOv2}},
	url = {http://arxiv.org/abs/2304.07193},
	doi = {10.48550/arXiv.2304.07193},
	urldate = {2025-11-23},
	publisher = {arXiv},
	author = {Oquab, Maxime and Darcet, Timothée and Moutakanni, Théo and Vo, Huy and Szafraniec, Marc and Khalidov, Vasil and Fernandez, Pierre and Haziza, Daniel and Massa, Francisco and El-Nouby, Alaaeldin and Assran, Mahmoud and Ballas, Nicolas and Galuba, Wojciech and Howes, Russell and Huang, Po-Yao and Li, Shang-Wen and Misra, Ishan and Rabbat, Michael and Sharma, Vasu and Synnaeve, Gabriel and Xu, Hu and Jegou, Hervé and Mairal, Julien and Labatut, Patrick and Joulin, Armand and Bojanowski, Piotr},
	month = feb,
	year = {2024},
	note = {arXiv:2304.07193 [cs]},
}

@misc{paiVisionFoundationModels2025,
	title = {Vision {Foundation} {Models} for {Computed} {Tomography}},
	url = {http://arxiv.org/abs/2501.09001},
	doi = {10.48550/arXiv.2501.09001},
	urldate = {2025-11-01},
	publisher = {arXiv},
	author = {Pai, Suraj and Hadzic, Ibrahim and Bontempi, Dennis and Bressem, Keno and Kann, Benjamin H. and Fedorov, Andriy and Mak, Raymond H. and Aerts, Hugo J. W. L.},
	month = feb,
	year = {2025},
	note = {arXiv:2501.09001 [eess]},
}

@article{wuGeneralistFoundationModel2025,
	title = {Towards generalist foundation model for radiology by leveraging web-scale {2D}\&{3D} medical data},
	volume = {16},
	copyright = {2025 The Author(s)},
	issn = {2041-1723},
	url = {https://www.nature.com/articles/s41467-025-62385-7},
	doi = {10.1038/s41467-025-62385-7},
	language = {en},
	number = {1},
	urldate = {2025-11-01},
	journal = {Nature Communications},
	publisher = {Nature Publishing Group},
	author = {Wu, Chaoyi and Zhang, Xiaoman and Zhang, Ya and Hui, Hui and Wang, Yanfeng and Xie, Weidi},
	month = aug,
	year = {2025},
	pages = {7866},
}

@article{zhuangMiMMaskMask2025,
	title = {{MiM}: {Mask} in {Mask} {Self}-{Supervised} {Pre}-{Training} for {3D} {Medical} {Image} {Analysis}},
	volume = {44},
	issn = {1558-254X},
	shorttitle = {{MiM}},
	url = {https://ieeexplore.ieee.org/document/10977020},
	doi = {10.1109/TMI.2025.3564382},
	number = {9},
	urldate = {2025-11-01},
	journal = {IEEE Transactions on Medical Imaging},
	author = {Zhuang, Jiaxin and Wu, Linshan and Wang, Qiong and Fei, Peng and Vardhanabhuti, Varut and Luo, Lin and Chen, Hao},
	month = sep,
	year = {2025},
	pages = {3727--3740},
}

@misc{idrist.MediastinalLymphNode2024,
	title = {Mediastinal {Lymph} {Node} {Quantification} ({LNQ})},
	doi = {10.7937/QVAZ-JA09},
	publisher = {The Cancer Imaging Archive},
	author = {{Idris, T.} and {Somarouthu, S.} and {Jacene, H.} and {LaCasce, A.} and {Ziegler, E.} and {Pieper, S.} and {Khajavi, R.} and {Dorent, R.} and {Pujol, S.} and {Kikinis, R.} and {Harris, G.}},
	year = {2024},
}

@misc{hanMycobacterialCTImaging2025,
	title = {Mycobacterial {CT} {Imaging} {Dataset}},
	doi = {10.34740/KAGGLE/DS/6248246},
	publisher = {Kaggle},
	author = {Han, Zhilin and Zhang, Yuyang and Ding, Wenlong and Xing, Zhiheng},
	year = {2025},
}

@misc{shahinOSICPulmonaryFibrosis2020,
	title = {{OSIC} {Pulmonary} {Fibrosis} {Progression}},
	url = {https://kaggle.com/competitions/osic-pulmonary-fibrosis-progression},
	publisher = {Kaggle},
	author = {Shahin, Ahmed and Wegworth, Carmela and {David} and Estes, Elizabeth and Elliott, Julia and Zita, Justin and Walsh, Simon and {Slepetys} and Cukierski, Will},
	year = {2020},
}

@misc{armatoiiiDataLIDCIDRI2015,
	title = {Data {From} {LIDC}-{IDRI}},
	copyright = {Creative Commons Attribution 3.0 Unported},
	url = {https://www.cancerimagingarchive.net/collection/lidc-idri/},
	doi = {10.7937/K9/TCIA.2015.LO9QL9SX},
	urldate = {2025-12-01},
	publisher = {The Cancer Imaging Archive},
	author = {Armato III, Samuel G. and McLennan, Geoffrey and Bidaut, Luc and McNitt-Gray, Michael F. and Meyer, Charles R. and Reeves, Anthony P. and Zhao, Binsheng and Aberle, Denise R. and Henschke, Claudia I. and Hoffman, Eric A. and Kazerooni, Ella A. and MacMahon, Heber and Van Beek, Edwin J.R. and Yankelevitz, David and Biancardi, Alberto M. and Bland, Peyton H. and Brown, Matthew S. and Engelmann, Roger M. and Laderach, Gary E. and Max, Daniel and Pais, Richard C. and Qing, David P.Y. and Roberts, Rachael Y. and Smith, Amanda R. and Starkey, Adam and Batra, Poonam and Caligiuri, Philip and Farooqi, Ali and Gladish, Gregory W. and Jude, C. Matilda and Munden, Reginald F. and Petkovska, Iva and Quint, Leslie E. and Schwartz, Lawrence H. and Sundaram, Baskaran and Dodd, Lori E. and Fenimore, Charles and Gur, David and Petrick, Nicholas and Freymann, John and Kirby, Justin and Hughes, Brian and Casteele, Alessi Vande and Gupte, Sangeeta and Sallam, Maha and Heath, Michael D. and Kuhn, Michael H. and Dharaiya, Ekta and Burns, Richard and Fryd, David S. and Salganicoff, Marcos and Anand, Vikram and Shreter, Uri and Vastagh, Stephen and Croft, Barbara Y. and Clarke, Laurence P.},
	collaborator = {{TCIA Team}},
	year = {2015},
}

@misc{boschHeadNeckCetuximab2015,
	title = {Head-{Neck} {Cetuximab}},
	copyright = {TCIA Limited Access License, Creative Commons Attribution 3.0 Unported},
	url = {https://www.cancerimagingarchive.net/collection/head-neck-cetuximab/},
	doi = {10.7937/K9/TCIA.2015.7AKGJUPZ},
	urldate = {2025-12-01},
	publisher = {The Cancer Imaging Archive},
	author = {Bosch, Walter R. and Straube, William L. and Matthews, John W. and Purdy, James A.},
	collaborator = {{TCIA Team}},
	year = {2015},
}

@misc{buattiCTRTSTRUCTRTDOSERTPLANSetsHead2024,
	title = {{CT}-{RTSTRUCT}-{RTDOSE}-{RTPLAN} {Sets} of {Head} and {Neck} {Cancers} {Treated} with {Identical} {Prescriptions} using {IMRT}: {An} {Open} {Dataset} for {Deep} {Learning} in {Treatment} {Planning}},
	copyright = {TCIA Limited Access License, Creative Commons Attribution 3.0 Unported},
	shorttitle = {{CT}-{RTSTRUCT}-{RTDOSE}-{RTPLAN} {Sets} of {Head} and {Neck} {Cancers} {Treated} with {Identical} {Prescriptions} using {IMRT}},
	url = {https://www.cancerimagingarchive.net/collection/hnc-imrt-70-33/},
	doi = {10.7937/AHQH-XC79},
	urldate = {2025-12-01},
	publisher = {The Cancer Imaging Archive},
	author = {Buatti, Jacob and Kabat, Christopher and Li, Ruiqi and Sivabhaskar, Sruthi and de Oliveira, Michelle and Papanikolaou, Nikos and Stathakis, Sotirios and Paragios, Nikos and Kirby, Neil},
	year = {2024},
}

@misc{liLargeScaleCTPET2020,
	title = {A {Large}-{Scale} {CT} and {PET}/{CT} {Dataset} for {Lung} {Cancer} {Diagnosis}},
	copyright = {Creative Commons Attribution 4.0 International},
	url = {https://www.cancerimagingarchive.net/collection/lung-pet-ct-dx/},
	doi = {10.7937/TCIA.2020.NNC2-0461},
	urldate = {2025-12-01},
	publisher = {The Cancer Imaging Archive},
	author = {Li, Ping and Wang, Shuo and Li, Tang and Lu, Jingfeng and HuangFu, Yunxin and Wang, Dongxue},
	year = {2020},
}

@misc{ericksonCancerGenomeAtlas2016,
	title = {The {Cancer} {Genome} {Atlas} {Uterine} {Corpus} {Endometrial} {Carcinoma} {Collection} ({TCGA}-{UCEC})},
	copyright = {Creative Commons Attribution 3.0 Unported},
	url = {https://www.cancerimagingarchive.net/collection/tcga-ucec/},
	doi = {10.7937/K9/TCIA.2016.GKJ0ZWAC},
	urldate = {2025-12-01},
	publisher = {The Cancer Imaging Archive},
	author = {Erickson, Bradley J. and Mutch, David and Lippmann, Lynne and Jarosz, Rose},
	collaborator = {{TCIA Team}},
	year = {2016},
}

@misc{beichelDataQINHEADNECK2015,
	title = {Data {From} {QIN}-{HEADNECK}},
	copyright = {TCIA Limited Access License, Creative Commons Attribution 3.0 Unported},
	url = {https://www.cancerimagingarchive.net/collection/qin-headneck/},
	doi = {10.7937/K9/TCIA.2015.K0F5CGLI},
	urldate = {2025-12-01},
	publisher = {The Cancer Imaging Archive},
	author = {Beichel, Reinhard R and Ulrich, Ethan J and Bauer, Christian and Wahle, A and Brown, B and Chang, T and Plichta, KA and Smith, BJ and Sunderland, JJ and Braun, T and Fedorov, Andrey and Clunie, David and Onken, M and Magnotta, Vincent A and Menda, Yusuf and Riesmeier, J and Pieper, Steve and Kikinis, Ron and Graham, MM and Casavant, Thomas and Sonka, M and Buatti, JM},
	collaborator = {Smith K, Clark K},
	year = {2015},
	note = {Version Number: 4},
}

@misc{muziDataRIDERLung2015,
	title = {Data {From} {RIDER} {Lung} {PET}-{CT}},
	copyright = {Creative Commons Attribution 3.0 Unported},
	url = {https://www.cancerimagingarchive.net/collection/rider-lung-pet-ct/},
	doi = {10.7937/K9/TCIA.2015.OFIP7TVM},
	urldate = {2025-12-01},
	publisher = {The Cancer Imaging Archive},
	author = {Muzi, Peter and Wanner, Michelle and Kinahan, Paul},
	collaborator = {{TCIA Team}},
	year = {2015},
}

@misc{saltzStonyBrookUniversity2021,
	title = {Stony {Brook} {University} {COVID}-19 {Positive} {Cases}},
	copyright = {Creative Commons Attribution 4.0 International},
	url = {https://www.cancerimagingarchive.net/collection/covid-19-ny-sbu/},
	doi = {10.7937/TCIA.BBAG-2923},
	urldate = {2025-12-01},
	publisher = {The Cancer Imaging Archive},
	author = {Saltz, Joel and Saltz, Mary and Prasanna, Prateek and Moffitt, Richard and Hajagos, Janos and Bremer, Erich and Balsamo, Joseph and Kurc, Tahsin},
	year = {2021},
}

@misc{nationalcancerinstituteclinicalproteomictumoranalysisconsortiumcptacClinicalProteomicTumor2018a,
	title = {The {Clinical} {Proteomic} {Tumor} {Analysis} {Consortium} {Clear} {Cell} {Renal} {Cell} {Carcinoma} {Collection} ({CPTAC}-{CCRCC})},
	copyright = {Creative Commons Attribution 3.0 Unported},
	url = {https://www.cancerimagingarchive.net/collection/cptac-ccrcc/},
	doi = {10.7937/K9/TCIA.2018.OBLAMN27},
	urldate = {2025-12-01},
	publisher = {The Cancer Imaging Archive},
	author = {{National Cancer Institute Clinical Proteomic Tumor Analysis Consortium (CPTAC)}},
	collaborator = {{TCIA Team}},
	year = {2018},
}

@misc{ericksonCancerGenomeAtlas2016a,
	title = {The {Cancer} {Genome} {Atlas} {Liver} {Hepatocellular} {Carcinoma} {Collection} ({TCGA}-{LIHC})},
	copyright = {Creative Commons Attribution 3.0 Unported},
	url = {https://www.cancerimagingarchive.net/collection/tcga-lihc/},
	doi = {10.7937/K9/TCIA.2016.IMMQW8UQ},
	urldate = {2025-12-01},
	publisher = {The Cancer Imaging Archive},
	author = {Erickson, Bradley J. and Kirk, Shanah and Lee, Yueh and Bathe, Oliver and Kearns, Melissa and Gerdes, Cindy and Rieger-Christ, Kimberly and Lemmerman, John},
	collaborator = {{TCIA Team}},
	year = {2016},
}

@misc{kirkCancerGenomeAtlas2016,
	title = {The {Cancer} {Genome} {Atlas} {Urothelial} {Bladder} {Carcinoma} {Collection} ({TCGA}-{BLCA})},
	copyright = {Creative Commons Attribution 3.0 Unported},
	url = {https://www.cancerimagingarchive.net/collection/tcga-blca/},
	doi = {10.7937/K9/TCIA.2016.8LNG8XDR},
	urldate = {2025-12-01},
	publisher = {The Cancer Imaging Archive},
	author = {Kirk, Shanah and Lee, Yueh and Lucchesi, Fabiano R. and Aredes, Natalia D. and Gruszauskas, Nicholas and Catto, James and Garcia, Kimberly and Jarosz, Rose and Duddalwar, Vinay and Varghese, Bino and Rieger-Christ, Kimberly and Lemmerman, John},
	collaborator = {{TCIA Team}},
	year = {2016},
}

@misc{moawadMultimodalityAnnotatedHCC2021,
	title = {Multimodality annotated {HCC} cases with and without advanced imaging segmentation},
	copyright = {Creative Commons Attribution 4.0 International},
	url = {https://www.cancerimagingarchive.net/collection/hcc-tace-seg/},
	doi = {10.7937/TCIA.5FNA-0924},
	urldate = {2025-12-01},
	publisher = {The Cancer Imaging Archive},
	author = {Moawad, Ahmed W. and Fuentes, David and Morshid, Ali and Khalaf, Ahmed M. and Elmohr, Mohab M. and Abusaif, Abdelrahman and Hazle, John D. and Kaseb, Ahmed O. and Hassan, Manal and Mahvash, Armeen and Szklaruk, Janio and Qayyom, Aliyya and Elsayes, Khaled},
	year = {2021},
}

@misc{lucchesiCancerGenomeAtlas2016,
	title = {The {Cancer} {Genome} {Atlas} {Stomach} {Adenocarcinoma} {Collection} ({TCGA}-{STAD})},
	copyright = {Creative Commons Attribution 3.0 Unported},
	url = {https://www.cancerimagingarchive.net/collection/tcga-stad/},
	doi = {10.7937/K9/TCIA.2016.GDHL9KIM},
	urldate = {2025-12-01},
	publisher = {The Cancer Imaging Archive},
	author = {Lucchesi, Fabiano R. and Aredes, Natália D.},
	collaborator = {{TCIA Team}},
	year = {2016},
}

@misc{hellerC4KCKiTSChallenge2019,
	title = {{C4KC} {KiTS} {Challenge} {Kidney} {Tumor} {Segmentation} {Dataset}},
	copyright = {Creative Commons Attribution 3.0 Unported},
	url = {https://www.cancerimagingarchive.net/collection/c4kc-kits/},
	doi = {10.7937/TCIA.2019.IX49E8NX},
	urldate = {2025-12-01},
	publisher = {The Cancer Imaging Archive},
	author = {Heller, Nicholas and Sathianathen, Niranjan and Kalapara, Arveen and Walczak, Edward and Moore, Keenan and Kaluzniak, Heather and Rosenberg, Joel and Blake, Paul and Rengel, Zachary and Oestreich, Makinna and Dean, Joshua and Tradewell, Michael and Shah, Aneri and Tejpaul, Resha and Edgerton, Zachary and Peterson, Matthew and Raza, Shaneabbas and Regmi, Subodh and Papanikolopoulos, Nikolaos and Weight, Christopher},
	collaborator = {{TCIA Team}},
	year = {2019},
}

@misc{madhaviDataAntiPD1Immunotherapy2019,
	title = {Data from {Anti}-{PD}-1 {Immunotherapy} {Lung}},
	doi = {10.7937/tcia.2019.zjjwb9ip},
	publisher = {The Cancer Imaging Archive},
	author = {Madhavi, P. and Patel, S. and Tsao, A. S.},
	year = {2019},
}

@article{hannaEffectShiftSchedule2018,
	title = {Effect of {Shift}, {Schedule}, and {Volume} on {Interpretive} {Accuracy}: {A}                    {Retrospective} {Analysis} of 2.9 {Million} {Radiologic} {Examinations}},
	volume = {287},
	issn = {0033-8419},
	shorttitle = {Effect of {Shift}, {Schedule}, and {Volume} on {Interpretive} {Accuracy}},
	url = {https://pubs.rsna.org/doi/abs/10.1148/radiol.2017170555},
	doi = {10.1148/radiol.2017170555},
	number = {1},
	urldate = {2025-11-19},
	journal = {Radiology},
	publisher = {Radiological Society of North America},
	author = {Hanna, Tarek N. and Lamoureux, Christine and Krupinski, Elizabeth A. and Weber, Scott and Johnson, Jamlik-Omari},
	month = apr,
	year = {2018},
	pages = {205--212},
}

@article{alexanderMandatingLimitsWorkload2022,
	title = {Mandating {Limits} on {Workload}, {Duty}, and {Speed} in                     {Radiology}},
	volume = {304},
	issn = {0033-8419},
	url = {https://pubs.rsna.org/doi/full/10.1148/radiol.212631},
	doi = {10.1148/radiol.212631},
	number = {2},
	urldate = {2025-11-18},
	journal = {Radiology},
	publisher = {Radiological Society of North America},
	author = {Alexander, Robert and Waite, Stephen and Bruno, Michael A. and Krupinski, Elizabeth A. and Berlin, Leonard and Macknik, Stephen and Martinez-Conde, Susana},
	month = aug,
	year = {2022},
	pages = {274--282},
}

@misc{baiM3DAdvancing3D2024,
	title = {{M3D}: {Advancing} {3D} {Medical} {Image} {Analysis} with {Multi}-{Modal} {Large} {Language} {Models}},
	shorttitle = {{M3D}},
	url = {http://arxiv.org/abs/2404.00578},
	doi = {10.48550/arXiv.2404.00578},
	urldate = {2025-11-01},
	publisher = {arXiv},
	author = {Bai, Fan and Du, Yuxin and Huang, Tiejun and Meng, Max Q.-H. and Zhao, Bo},
	month = mar,
	year = {2024},
	note = {arXiv:2404.00578 [cs]},
}

@misc{zolotovaBurdenkosGlioblastomaProgression2023,
	title = {Burdenko's {Glioblastoma} {Progression} {Dataset} ({Burdenko}-{GBM}-{Progression})},
	copyright = {TCIA Limited Access License, Creative Commons Attribution 4.0 International},
	url = {https://www.cancerimagingarchive.net/collection/burdenko-gbm-progression/},
	doi = {10.7937/E1QP-D183},
	urldate = {2025-12-01},
	publisher = {The Cancer Imaging Archive},
	author = {Zolotova, Svetlana V. and Golanov, Andrey V. and Pronin, Igor N. and Dalechina, Alexandra V. and Nikolaeva, Anna A. and Belyashova, Alexandra S. and Usachev, Dmitry Y. and Kondrateva, Ekaterina A. and Druzhinina, Polina V. and Shirokikh, Boris N. and Saparov, Talgat N. and Belyaev, Mikhail G. and Kurmukov, Anvar I.},
	year = {2023},
}

@misc{shusharinaGliomaImageSegmentation2021,
	title = {Glioma {Image} {Segmentation} for {Radiotherapy}: {RT} targets, barriers to cancer spread, and organs at risk ({GLIS}-{RT})},
	copyright = {TCIA Limited Access License},
	shorttitle = {Glioma {Image} {Segmentation} for {Radiotherapy}},
	url = {https://www.cancerimagingarchive.net/collection/glis-rt/},
	doi = {10.7937/TCIA.T905-ZQ20},
	urldate = {2025-12-01},
	publisher = {The Cancer Imaging Archive},
	author = {Shusharina, Nadya and Bortfeld, Thomas},
	year = {2021},
}

@misc{zuleyCancerGenomeAtlas2016,
	title = {The {Cancer} {Genome} {Atlas} {Head}-{Neck} {Squamous} {Cell} {Carcinoma} {Collection} ({TCGA}-{HNSC})},
	copyright = {TCIA Limited Access License},
	url = {https://www.cancerimagingarchive.net/collection/tcga-hnsc/},
	doi = {10.7937/K9/TCIA.2016.LXKQ47MS},
	urldate = {2025-12-01},
	publisher = {The Cancer Imaging Archive},
	author = {Zuley, Margarita L. and Jarosz, Rose and Kirk, Shanah and Lee, Yueh and Colen, Rivka and Garcia, Kimberly and Delbeke, Dominique and Pham, Michelle and Nagy, Paul and Sevinc, Gorkem and Goldsmith, Marla and Khan, Subair and Net, Jose M. and Lucchesi, Fabiano R. and Aredes, Natalia D.},
	collaborator = {{TCIA Team}},
	year = {2016},
}

@misc{vallieresDataHeadNeckPETCT2017,
	title = {Data from {Head}-{Neck}-{PET}-{CT}},
	copyright = {TCIA Limited Access License, Creative Commons Attribution 3.0 Unported},
	url = {https://www.cancerimagingarchive.net/collection/head-neck-pet-ct/},
	doi = {10.7937/K9/TCIA.2017.8OJE5Q00},
	urldate = {2025-12-01},
	publisher = {The Cancer Imaging Archive},
	author = {Vallières, Martin and Kay-Rivest, Emily and Perrin, Léo and Liem, Xavier and Furstoss, Christophe and Khaouam, Nader and Nguyen-Tan, Phuc and Wang, Chang-Shu and Sultanem, Khalil},
	collaborator = {{TCIA Team}},
	year = {2017},
}

@misc{smithDataCTCOLONOGRAPHY2015,
	title = {Data {From} {CT} {COLONOGRAPHY}},
	doi = {10.7937/K9/TCIA.2015.NWTESAY1},
	publisher = {The Cancer Imaging Archive},
	author = {Smith, K. and Clark, K. and Bennett, W. and Nolan, T. and Kirby, J. and Wolfsberger, M. and Moulton, J. and Vendt, B. and Freymann, J.},
	year = {2015},
}

@misc{weeDataHEADNECKRADIOMICSHN12019,
	title = {Data from {HEAD}-{NECK}-{RADIOMICS}-{HN1}},
	doi = {10.7937/tcia.2019.8kap372n},
	publisher = {The Cancer Imaging Archive},
	author = {Wee, L. and Dekker, A.},
	year = {2019},
}

@misc{tongAbdominalPelvicEnhanced2022,
	title = {Abdominal or pelvic enhanced {CT} images within 10 days before surgery of 230 patients with stage {II} colorectal cancer ({StageII}-{Colorectal}-{CT})},
	doi = {10.7937/p5k5-tg43},
	publisher = {The Cancer Imaging Archive},
	author = {Tong, T. and Li, M.},
	year = {2022},
}

@misc{nationalcancerinstituteclinicalproteomictumoranalysisconsortiumcptacClinicalProteomicTumor2018,
	title = {The {Clinical} {Proteomic} {Tumor} {Analysis} {Consortium} {Head} and {Neck} {Squamous} {Cell} {Carcinoma} {Collection} ({CPTAC}-{HNSCC}) ({Version} 19)},
	doi = {10.7937/k9/tcia.2018.uw45nh81},
	publisher = {The Cancer Imaging Archive},
	collaborator = {{National Cancer Institute Clinical Proteomic Tumor Analysis Consortium (CPTAC)}},
	year = {2018},
}

@article{welchRADCUREOpensourceHead2024,
	title = {{RADCURE}: {An} open-source head and neck cancer {CT} dataset for clinical radiation therapy insights},
	volume = {51},
	issn = {2473-4209},
	shorttitle = {{RADCURE}},
	doi = {10.1002/mp.16972},
	language = {eng},
	number = {4},
	journal = {Medical Physics},
	author = {Welch, Mattea L. and Kim, Sejin and Hope, Andrew J. and Huang, Shao Hui and Lu, Zhibin and Marsilla, Joseph and Kazmierski, Michal and Rey-McIntyre, Katrina and Patel, Tirth and O'Sullivan, Brian and Waldron, John and Bratman, Scott and Haibe-Kains, Benjamin and Tadic, Tony and {Princess Margaret Head and Neck Site Group}},
	month = apr,
	year = {2024},
	pages = {3101--3109},
}

@article{colakRSNAPulmonaryEmbolism2021,
	title = {The {RSNA} {Pulmonary} {Embolism} {CT} {Dataset}},
	volume = {3},
	url = {https://pubs.rsna.org/doi/full/10.1148/ryai.2021200254},
	doi = {10.1148/ryai.2021200254},
	number = {2},
	urldate = {2025-11-28},
	journal = {Radiology: Artificial Intelligence},
	publisher = {Radiological Society of North America},
	author = {Colak, Errol and Kitamura, Felipe C. and Hobbs, Stephen B. and Wu, Carol C. and Lungren, Matthew P. and Prevedello, Luciano M. and Kalpathy-Cramer, Jayashree and Ball, Robyn L. and Shih, George and Stein, Anouk and Halabi, Safwan S. and Altinmakas, Emre and Law, Meng and Kumar, Parveen and Manzalawi, Karam A. and Nelson Rubio, Dennis Charles and Sechrist, Jacob W. and Germaine, Pauline and Lopez, Eva Castro and Amerio, Tomas and Gupta, Pushpender and Jain, Manoj and Kay, Fernando U. and Lin, Cheng Ting and Sen, Saugata and Revels, Jonathan Wesley and Brussaard, Carola C. and Mongan, John and {For the RSNA-STR Annotators and Dataset Curation Contributors}},
	month = mar,
	year = {2021},
	pages = {e200254},
}

@misc{kinahanDataACRIN6685,
	title = {Data from the {ACRIN} 6685 {Trial} {HNSCC}-{FDG}-{PET}/{CT}},
	doi = {10.7937/K9/TCIA.2016.JQEJZZNG},
	publisher = {TCIA},
	author = {Kinahan, P. and Muzi, M. and Bialecki, B. and Coombs, L.},
}

@misc{grossbergHNSCCVersion42020,
	title = {{HNSCC} {Version} 4},
	doi = {10.7937/k9/tcia.2020.a8sh-7363},
	publisher = {The Cancer Imaging Archive},
	author = {Grossberg, A and Elhalawani, H and Mohamed, A and Mulder, S and Williams, B and White, A L and Zafereo, J and Wong, A J and Berends, J E and AboHashem, S and Aymard, J M and Kanwar, A and Perni, S and Rock, C D and Chamchod, S and Kantor, M and Browne, T and Hutcheson, K and Gunn, G B and Frank, S J and Rosenthal, D I and Garden, A S and Fuller, C D},
	collaborator = {{M.D. Anderson Cancer Center Head and Neck Quantitative Imaging Working Group}},
	year = {2020},
}

@article{rudieRSNAAbdominalTraumatic2024,
	title = {The {RSNA} {Abdominal} {Traumatic} {Injury} {CT} ({RATIC})                     {Dataset}},
	volume = {6},
	url = {https://pubs.rsna.org/doi/10.1148/ryai.240101},
	doi = {10.1148/ryai.240101},
	number = {6},
	urldate = {2025-11-28},
	journal = {Radiology: Artificial Intelligence},
	publisher = {Radiological Society of North America},
	author = {Rudie, Jeffrey D. and Lin, Hui-Ming and Ball, Robyn L. and Jalal, Sabeena and Prevedello, Luciano M. and Nicolaou, Savvas and Marinelli, Brett S. and Flanders, Adam E. and Magudia, Kirti and Shih, George and Davis, Melissa A. and Mongan, John and Chang, Peter D. and Berger, Ferco H. and Hermans, Sebastiaan and Law, Meng and Richards, Tyler and Grunz, Jan-Peter and Kunz, Andreas Steven and Mathur, Shobhit and Galea-Soler, Sandro and Chung, Andrew D. and Afat, Saif and Kuo, Chin-Chi and Aweidah, Layal and Villanueva Campos, Ana and Somasundaram, Arjuna and Sanchez Tijmes, Felipe Antonio and Jantarangkoon, Attaporn and Kayat Bittencourt, Leonardo and Brassil, Michael and El Hajjami, Ayoub and Dogan, Hakan and Becircic, Muris and Bharatkumar, Agrahara G. and Júdice de Mattos Farina, Eduardo Moreno and Colak, Errol and {for the Dataset Curator Group} and {Dataset Contributor Group} and {Dataset Annotator Group}},
	month = nov,
	year = {2024},
	pages = {e240101},
}

@article{wasserthalTotalSegmentatorRobustSegmentation2023,
	title = {{TotalSegmentator}: {Robust} {Segmentation} of 104 {Anatomic} {Structures} in {CT} {Images}},
	volume = {5},
	shorttitle = {{TotalSegmentator}},
	url = {https://pubs.rsna.org/doi/10.1148/ryai.230024},
	doi = {10.1148/ryai.230024},
	number = {5},
	urldate = {2025-11-28},
	journal = {Radiology: Artificial Intelligence},
	publisher = {Radiological Society of North America},
	author = {Wasserthal, Jakob and Breit, Hanns-Christian and Meyer, Manfred T. and Pradella, Maurice and Hinck, Daniel and Sauter, Alexander W. and Heye, Tobias and Boll, Daniel T. and Cyriac, Joshy and Yang, Shan and Bach, Michael and Segeroth, Martin},
	month = sep,
	year = {2023},
	pages = {e230024},
}

@article{maAbdomenCT1KAbdominalOrgan2022,
	title = {{AbdomenCT}-{1K}: {Is} {Abdominal} {Organ} {Segmentation} a {Solved} {Problem}?},
	volume = {44},
	issn = {1939-3539},
	shorttitle = {{AbdomenCT}-{1K}},
	url = {https://ieeexplore.ieee.org/document/9497733},
	doi = {10.1109/TPAMI.2021.3100536},
	number = {10},
	urldate = {2025-11-28},
	journal = {IEEE Transactions on Pattern Analysis and Machine Intelligence},
	author = {Ma, Jun and Zhang, Yao and Gu, Song and Zhu, Cheng and Ge, Cheng and Zhang, Yichi and An, Xingle and Wang, Congcong and Wang, Qiyuan and Liu, Xin and Cao, Shucheng and Zhang, Qi and Liu, Shangqing and Wang, Yunpeng and Li, Yuhui and He, Jian and Yang, Xiaoping},
	month = oct,
	year = {2022},
	pages = {6695--6714},
}

@misc{nationallungscreeningtrialresearchteamDataNationalLung2013,
	title = {Data from the {National} {Lung} {Screening} {Trial} ({NLST})},
	doi = {10.7937/TCIA.HMQ8-J677},
	publisher = {The Cancer Imaging Archive},
	author = {{National Lung Screening Trial Research Team}},
	year = {2013},
}

@misc{huangINSPECTMultimodalDataset2023,
	title = {{INSPECT}: {A} {Multimodal} {Dataset} for {Pulmonary} {Embolism} {Diagnosis} and {Prognosis}},
	shorttitle = {{INSPECT}},
	url = {http://arxiv.org/abs/2311.10798},
	doi = {10.48550/arXiv.2311.10798},
	urldate = {2025-11-28},
	publisher = {arXiv},
	author = {Huang, Shih-Cheng and Huo, Zepeng and Steinberg, Ethan and Chiang, Chia-Chun and Lungren, Matthew P. and Langlotz, Curtis P. and Yeung, Serena and Shah, Nigam H. and Fries, Jason A.},
	month = nov,
	year = {2023},
	note = {arXiv:2311.10798 [cs]},
}

\end{document}